\documentclass[sigconf,screen,nonacm]{acmart}

\usepackage{multirow}
\usepackage{booktabs}
\usepackage{bm}
\usepackage{enumitem}
\usepackage{algorithm}
\usepackage{algpseudocode}
\usepackage{verbatim}
\usepackage{hyperref}

\AtBeginDocument{%
	\providecommand\BibTeX{{%
			\normalfont B\kern-0.5em{\scshape i\kern-0.25em b}\kern-0.8em\TeX}}}

\copyrightyear{2022}
\acmYear{2022}
\setcopyright{acmcopyright}
\acmConference[MM '22]{Proceedings of the 30th ACM International Conference on Multimedia}{October 10--14, 2022}{Lisboa, Portugal}
\acmBooktitle{Proceedings of the 30th ACM International Conference on Multimedia (MM '22), October 10--14, 2022, Lisboa, Portugal}
\acmPrice{15.00}
\acmDOI{XXXXXXX.XXXXXXX}
\acmISBN{978-1-4503-XXXX-X/18/06}

\begin{document}
	
	\title{Diverse Human Motion Prediction via Gumbel-Softmax Sampling from an Auxiliary Space}
	
	
	\author{Lingwei Dang}
	\affiliation{%
		\institution{South China University of Technology}
		\city{Guangzhou}
		\state{Guangdong}
		\country{China}}
	\email{csdanglw@mail.scut.edu.cn}
	
	\author{Yongwei Nie}
	\authornote{Corresponding author.}
	\affiliation{%
		\institution{South China University of Technology}
		\city{Guangzhou}
		\state{Guangdong}
		\country{China}}
	\email{nieyongwei@scut.edu.cn}
	
	\author{Chengjiang Long}
	\affiliation{%
		\institution{Meta Reality Lab}
		\city{Burlingame}
		\state{CA}
		\country{USA}}
	\email{clong1@fb.com}
	
	\author{Qing Zhang}
	\affiliation{%
		\institution{Sun Yat-sen University}
		\city{Guangzhou}
		\state{Guangdong}
		\country{China}}
	\email{zhangqing.whu.cs@gmail.com}
	
	\author{Guiqing Li}
	\affiliation{%
		\institution{South China University of Technology}
		\city{Guangzhou}
		\state{Guangdong}
		\country{China}}
	\email{ligq@scut.edu.cn}
	
	\begin{abstract}
		Diverse human motion prediction aims at predicting multiple possible future pose sequences from a sequence of observed poses. Previous approaches usually employ deep generative networks to model the conditional distribution of data, and then randomly sample outcomes from the distribution. While different results can be obtained, they are usually the most likely ones which are not diverse enough. Recent work explicitly learns multiple modes of the conditional distribution via a deterministic network, which however can only cover a fixed number of modes within a limited range. In this paper, we propose a novel sampling strategy for sampling very diverse results from an imbalanced multimodal distribution learned by a deep generative model. Our method works by generating an auxiliary space and smartly making randomly sampling from the auxiliary space equivalent to the diverse sampling from the target distribution. We propose a simple yet effective network architecture that implements this novel sampling strategy, which incorporates a Gumbel-Softmax coefficient matrix sampling method and an aggressive diversity promoting hinge loss function. Extensive experiments demonstrate that our method significantly improves both the diversity and accuracy of the samplings compared with previous state-of-the-art sampling approaches. Code and pre-trained models are available at \href{https://github.com/Droliven/diverse_sampling}{https://github.com/Droliven/diverse\_sampling}.
	\end{abstract}

	
	

	\begin{teaserfigure}
		\centering
		\includegraphics[width=1.0\textwidth]{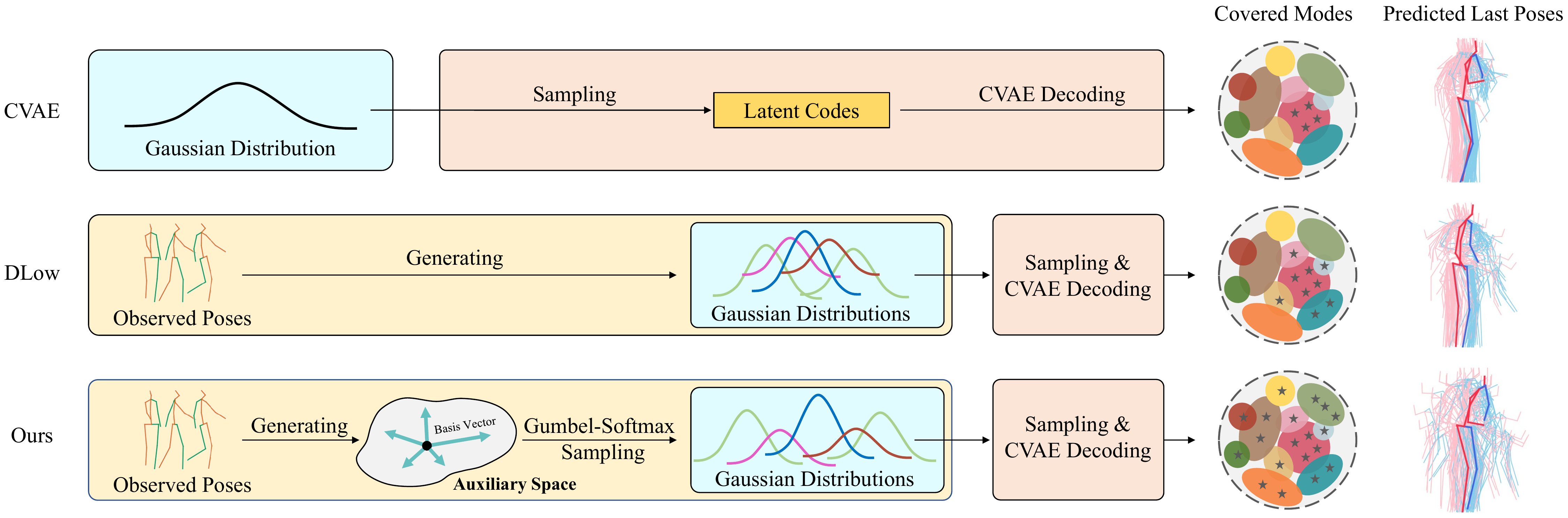} 
		\caption{Different strategies for sampling diverse results from an imbalanced multimodal distribution. The vanilla CVAE model randomly samples latent codes from a prior distribution which are then decoded into results that only reside in the major mode of the target distribution. DLow~\cite{yuan2020dlow} first generates multiple Gaussian distributions, and then samples latent codes from different Gaussian priors. The Gaussian priors can be viewed as corresponding to different modes of the target distribution, therefore this method can cover more modes than random sampling. Our method generates multiple Gaussian distributions by sampling points from an auxiliary space. Due to the high flexibility and capacity of the space, our method is able to cover even more modes than DLow. The rightmost are the last poses of future pose sequences predicted from a given input, all stacked together to visually show that our results are more diverse than the others.}
		\label{fig:teaser}
	\end{teaserfigure}

	\maketitle
	
	\section{Introduction}
	Human Motion Prediction (HMP) has a wide range of applications in autonomous driving, human-robot interaction, and animation creation. Most previous works \cite{martinez2017human, song2017end, sang2020human, chiu2019action, corona2020context, aksan2019structured, li2017auto, fragkiadaki2015recurrent, liu2022investigating,mao2019learning, mao2020history, li2020dynamic, li2021symbiotic, cui2020learning, dang2021msr, cui2021towards, li2020multitask, liu2021motion, aksan2021spatio, martinez2021pose, aksan2020attention, cai2020learning, su2021motion, Ma_2022_CVPR} perform deterministic HMP that only generates one result in the future. Recently, many diverse HMP approaches can predict multiple possible future motions. Due to the stochasticity of human motion, multiple solutions naturally exist, and forecasting them is of great importance in practice. For example, it would be better for a vehicle to know that a pedestrian in front of it may not only walk ahead but also turn left suddenly. 
	
	Diverse HMP approaches like \cite{barsoum2018hp,kundu2019bihmp,yan2018mt} adopt deep generative networks, such as GAN~\cite{goodfellow2014generative} or CVAE~\cite{kingma2013auto}, to learn a conditional distribution of future poses given previous ones. Taking CVAE as an example (see Figure~\ref{fig:teaser} top), after training a CVAE, one can randomly sample latent codes (noises) from a prior distribution (\textit{e.g.}, a Gaussian prior), and then decode the random noises to future sequences by the CVAE decoder. However, since the CVAE model is obtained by maximizing the likelihood of the training data that is often highly imbalanced, it usually learns an imbalanced multimodal conditional distribution. Latent codes drawn at random from the prior distribution most probably correspond to the most likely results that fall in the dominant mode of the distribution of data, while ignoring other results of low probability but high fidelity.
	
	Recently, Yuan \textit{et al.} \cite{yuan2020dlow} proposed a method called DLow sampling. As shown in the second row of Figure~\ref{fig:teaser}, given observed poses, DLow uses a neural network to generate multiple Gaussian distributions, and then samples latent codes from all the generated Gaussian distributions. They optimize the network to diversify the Gaussian distributions, making them corresponding to different modes of the target distribution. However, directly generating Gaussian distributions has two limitations. Firstly, a network can only generate a fixed number of Gaussian distributions, while there may exist much more modes in the target data distribution. Secondly, it entangles the performance of diverse prediction with the learning of the network parameters, requiring the latter to consider all training data and make tradeoffs between them, thus in turn limiting the diverse prediction performance.
	
	In this paper, we propose a sampling strategy that disentangles the above direct dependency between a network and the intermediate Gaussian distributions. As shown in the third row of Figure~\ref{fig:teaser}, instead of Gaussian distributions, we learn a set of basis vectors from the observed poses. We assume that the basis vectors determine an auxiliary space, and any linear combination of the basis vectors corresponds to a point in the auxiliary space. We randomly sample a set of points from the auxiliary space by the Gumbel-Softmax sampling strategy, and then map them to Gaussian distributions which finally correspond to different modes of the target distribution. In other words, we use a network to learn an auxiliary space, and build the following connection between the auxiliary space and the target distribution: \textit{randomly sampling from the auxiliary space corresponds to diverse sampling from the target distribution}. The diverse prediction is now tied to the structure of the auxiliary space rather than directly to the parameters of a network. Since the auxiliary space can be flexibly deformed in terms of both size and shape, our sampling method can cover all modes of the target distribution in theory, supporting very diverse human motion prediction. Note that at the training stage, we sample a fixed number of points from the auxiliary space, which facilitates the training of the auxiliary space. After training, since the shape of the auxiliary space has already been constructed, we can sample any number of points from it.
	
	The Gumbel-Softmax sampling method samples points from the auxiliary space by generating a coefficient matrix that linearly combines the basis vectors. There exist other sampling strategies such as Uniform-Softmax sampling and Gaussian-Softmax sampling. We compare with them and find that the Gumbel-Softmax sampling is more effective as it is more aggressive in assigning larger weights to relatively fewer basis vectors. Training/testing with these weights can make better use of each basis vector to sample more distinctive points from the auxiliary space that correspond to more diverse modes of the target distribution. Finally, In order to train our model, we propose a hinge-diversity loss function which explicitly requires the distance between any pair of predictions to be greater than a user-specified threshold. The hinge-diversity loss further strengthens the diversity of predictions while less affecting their accuracy.
	
	In summary, the contributions of this work are three-fold:
	\begin{itemize}
		\item We propose a novel sampling method that is highly capable and convenient for diverse and accurate sampling from a complex imbalanced multimodal distribution, by converting sampling from the distribution into randomly sampling of points from an auxiliary space.
		\item We propose a Gumbel-Softmax sampling method to sample points from the auxiliary space, and a hinge-diversity loss to train our framework, both of which further improve the performance of our method.
		\item Extensive comparisons and ablation experimental results conducted on Human3.6M \cite{ionescu2013human3} and HumanEva-I \cite{sigal2010humaneva} demonstrate the effectiveness of our approach.
	\end{itemize}

	\section{Related Work}
	
	\textbf{Deterministic Human Motion Prediction.} Most previous approaches target deterministic HMP, by which only one output is produced per sequence of historical poses. Considering the ability of Recurrent Neural Networks (RNNs) in modeling temporal dependencies of sequential data, many approaches \cite{martinez2017human, song2017end, sang2020human, chiu2019action, corona2020context, aksan2019structured, li2017auto, fragkiadaki2015recurrent, liu2022investigating} use RNNs to tackle the sequence-to-sequence HMP problem, which however usually suffer from problems of discontinuity and error accumulation. Instead of RNNs, recent works \cite{mao2019learning, mao2020history, li2020dynamic, li2021symbiotic, cui2020learning, dang2021msr, cui2021towards, li2020multitask, liu2021motion, mao2021multi, Ma_2022_CVPR} employ Graph Convolutional Networks (GCNs)~\cite{Shi:CVPR2021, Duan:AAAI2022, Shi:AAAI2022} for this task, as GCNs are effective in discovering spatial and temporal relations between pairs of human joints. Similar to GCNs, Transformer \cite{vaswani2017attention, Dong:MM2021} can capture long-term dependencies between human joints, and has been adapted to handle the deterministic HMP problem \cite{cai2020learning, aksan2021spatio, martinez2021pose, aksan2020attention}. Different from the above methods, this paper attempts to tackle stochastic HMP which outputs multiple possible results given one input.
	
	\textbf{Diverse Human Motion Prediction.} Many efforts have been paid to the diverse HMP problem \cite{barsoum2018hp,yan2018mt,kundu2019bihmp,yuan2020dlow,aliakbarian2020stochastic,liu2021aggregated,tanke2021intention,lyu2021learning,mao2021generating,aliakbarian2021contextually,cai2021unified}. For example, Barsoum \textit{et al.} \cite{barsoum2018hp} proposed HP-GAN which is a generative adversarial framework that models the probability density function of future human poses conditioned on given poses. At test time, a random vector $\mathbf{z}$ controls the generation of different future poses. Yan \textit{et al.} \cite{yan2018mt} proposed MT-VAE using VAE \cite{kingma2013auto, Liu:ICCV2021} to model the conditional distribution of data. In MT-VAE, a random variable $\mathbf{z}$ encodes a latent transformation that transforms the observed poses to specific future poses. Both GANs and VAEs randomly sample latent vectors $\mathbf{z}$ from a prior distribution which however are usually decoded into similar results. To alleviate the problem, Kundu \textit{et al.} \cite{kundu2019bihmp} proposed BiHMP-GAN in which a discriminator is used to regress the random vector $\mathbf{z}$ originally fed into the generator, enforcing one-to-one mapping between the latent vector $\mathbf{z}$ and the corresponding motion prediction. Aliakbarian \textit{et al.} \cite{aliakbarian2020stochastic} believed that the generative models tend to ignore the random vectors. To prevent such ignoring, they proposed a Mix-and-Match perturbation mechanism to sufficiently mix random noises and conditional poses in \cite{aliakbarian2020stochastic}. In their later work \cite{aliakbarian2021contextually}, a random noise is generated directly conditioned on the input poses. Instead of randomly sampling $\mathbf{z}$, Yuan \textit{et al.} \cite{yuan2020dlow} proposed a sampling strategy called DLow by which different random vectors that correspond to diverse predictions are explicitly inferred, achieving impressive results in sampling from minor modes. However, DLow is limited by its design of inferring random vectors directly from a network. Our method disentangles this dependency and obtains more diverse results with higher accuracy. Recently, the method of \cite{mao2021generating} directly maps a random vector together with the observed poses to a future sequence, without relying on generative models. For results of different random vectors but the same input poses, it applies a diversity loss to enlarge the differences between them, and meanwhile uses many prior constraints to guarantee their plausibility. We compare with this very different method and show that our method outperforms it on diversity and accuracy metrics.

	\section{Methodology}
	\begin{figure*}[!t]
		\centering
		\includegraphics[width=1.0\linewidth]{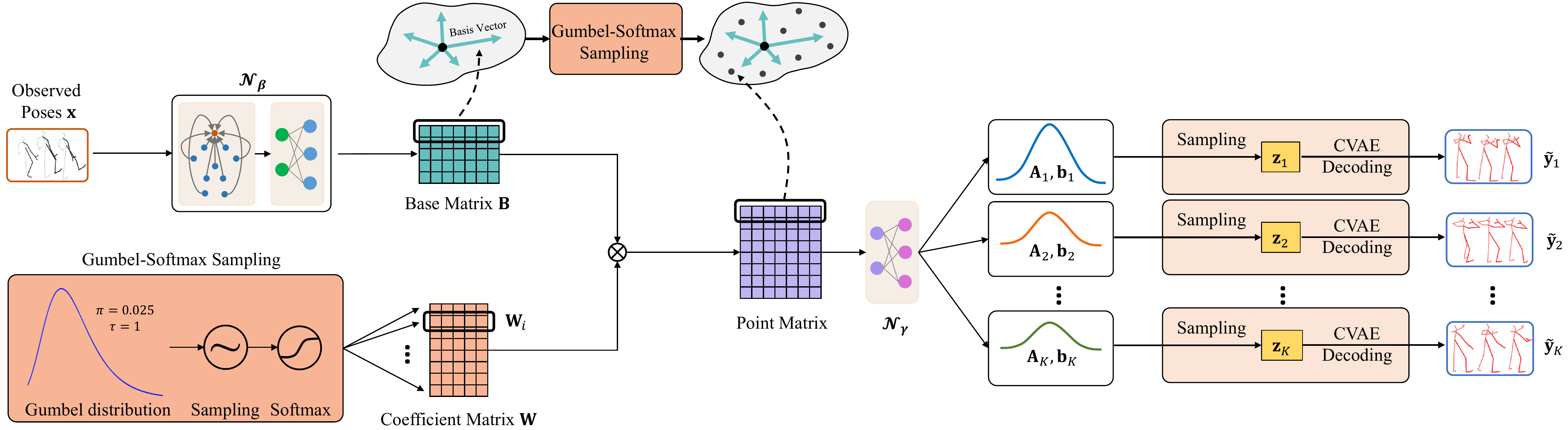}
		\caption{On one hand, we use a network $\mathcal{N}_{\bm{\beta}}$ to generate a base matrix from the observed poses. On the other hand, we employ the Gumbel-Softmax sampling method to generate a coefficient matrix. The multiplication of the two matrices samples multiple points from the auxiliary space determined by the base matrix. We then employ another network $\mathcal{N}_{\bm{\gamma}}$ to map these points to a set of Gaussian distributions from which latent codes are drawn and finally decoded into future pose sequences.}
		\label{fig:pipeline}
	\end{figure*}
	
	We use CVAE to model the distribution of data, then propose a post-hoc sampling strategy to sample diverse results from the distribution. We therefore first introduce the background of CVAE-based stochastic HMP. Since our method is based on DLow~\cite{yuan2020dlow}, we also briefly introduce DLow, and finally describe our method in detail.
	
	\subsection{Background}
	
	\subsubsection{CVAE-based Stochastic Prediction}
	Let $p(\mathbf{y}|\mathbf{x})$ denote the distribution of $\mathbf{y}$ given $\mathbf{x}$, where $\mathbf{x}$ is an observed pose sequence and $\mathbf{y}$ is a possible future pose sequence that may appear after $\mathbf{x}$. To sample $\mathbf{y}$ from $p(\mathbf{y}|\mathbf{x})$, one usually introduces a latent variable $\mathbf{z}$ and reparameterizes $p(\mathbf{y}|\mathbf{x})$ as $p(\mathbf{y}|\mathbf{x})=\int p(\mathbf{y}|\mathbf{x},\mathbf{z})p(\mathbf{z})d\mathbf{z}$. Then, $\mathbf{y}$ can be generated in two steps:
	\begin{equation}
		\label{eq:random-sampling}
		\begin{aligned}
			\mathbf{z} & \sim p(\mathbf{z}), \\
			\mathbf{y} & = \mathcal{G}_{\bm{\theta}}(\mathbf{x},\mathbf{z}),
		\end{aligned}
	\end{equation}
	where a random vector $\mathbf{z}$ is sampled from a prior distribution $p(\mathbf{z})$ (\textit{e.g.}, Gaussian) at first, then $\mathbf{y}$ is generated by a deterministic function $\mathcal{G}_{\bm{\theta}}$ parameterized by $\bm{\theta}$ taking $\mathbf{x}$ and $\mathbf{z}$ as input. To learn $\mathcal{G}_{\bm{\theta}}$, a popular way is to use a CVAE which maximizes the log-likelihood of data $\mathbf{y}$ given $\mathbf{x}$, by introducing an approximate posterior $q(\mathbf{z}|\mathbf{x},\mathbf{y})$ and maximizing the following evidence lower bound (ELBO):
	\begin{equation}
		\begin{aligned}
			\log p(\mathbf{y}|\mathbf{x}) =\log & \int p(\mathbf{y}|\mathbf{x},\mathbf{z})p(\mathbf{z})d\mathbf{z} \\
			=\log \int  \frac{p(\mathbf{y}|\mathbf{x},\mathbf{z})p(\mathbf{z})}{q(\mathbf{z}|\mathbf{x},\mathbf{y})} q(\mathbf{z}|\mathbf{x},\mathbf{y})& d\mathbf{z}
			\geq \mathbb{E}_{q} \log \frac{p(\mathbf{y}|\mathbf{x},\mathbf{z})p(\mathbf{z})}{q(\mathbf{z}|\mathbf{x},\mathbf{y})}.
		\end{aligned}
	\end{equation}
	CVAE models the two distributions of $q(\mathbf{z}|\mathbf{x},\mathbf{y})$ and $p(\mathbf{y}|\mathbf{x},\mathbf{z})$ by two neural networks $\mathcal{F}_{\bm{\phi}}$ and $\mathcal{G}_{\bm{\theta}}$, and estimates their parameters by optimizing the following loss function:
	\begin{equation}
		\label{eq:elbo-loss}
		\mathcal{L}(\mathbf{y}, \bm{\phi}, \bm{\theta}) = -\mathcal{KL}\left(q_{\bm{\phi}}(\mathbf{z}|\mathbf{x},\mathbf{y})\| p(\mathbf{z})\right) + \mathbb{E}_{q_{\bm{\phi}}} \log p_{\bm{\theta}}(\mathbf{y}|\mathbf{x},\mathbf{z}).
	\end{equation}
	During training, the encoder $\mathcal{F}_{\bm{\phi}}$ first generates $\mathbf{z}$ given $\mathbf{x}$ and $\mathbf{y}$, and then the decoder $\mathcal{G}_{\bm{\theta}}$ reconstructs the input $\mathbf{y}$ given $\mathbf{z}$ and $\mathbf{x}$. 
	At test time, one can sample a $\mathbf{z}$ from the prior distribution $p(\mathbf{z})$, and then predict a $\mathbf{\tilde{y}}$ by $\mathcal{G}_{\bm{\theta}}$ given $\mathbf{z}$ and $\mathbf{x}$. For multiple predictions, one needs to sample $\mathbf{z}_1,\cdots,\mathbf{z}_K$ independently, and predict $\mathbf{\tilde{y}}_1,\cdots,\mathbf{\tilde{y}}_K$ accordingly using the same $\mathbf{x}$. However, extensive experiments demonstrate that the diversity of $\mathbf{\tilde{y}}_1,\cdots,\mathbf{\tilde{y}}_K$ is not satisfactory. 

	\subsubsection{DLow Sampling}
	To enable diverse prediction, Yuan \textit{et al.} \cite{yuan2020dlow} proposed DLow. Given $\mathbf{x}$, they used a network $\mathcal{Q}_{\bm{\psi}}$ parameterized by $\bm{\psi}$ to generate $K$ Gaussian distributions: 
	\begin{equation}
		\{(\mathbf{A}_k,\mathbf{b}_k)\}_{k=1}^{K} = \mathcal{Q}_{\bm{\psi}}(\mathbf{x}),
	\end{equation}
	where $\mathbf{A}_k\in \mathbb{R}^{n_z\times n_z}$, $\mathbf{b}_k \in \mathbb{R}^{n_z}$ are variance and mean of the $k^{th}$ Gaussian distribution, and $n_z$ is the dimension size. Then, they predicted $K$ results by:
	\begin{equation}
		\label{eq:dlow-sampling}
		\begin{aligned}
			&\bm{\epsilon}  \sim \mathcal{N}(0,1),\\ 
			&\mathbf{z}_k  = \mathbf{A}_k \mathbf{\bm{\epsilon}} + \mathbf{b}_k, &1\leq k \leq K,\\
			&\mathbf{y}_k  = \mathcal{G}_{\bm{\theta}}(\mathbf{x},\mathbf{z}_k), &1\leq k \leq K.
		\end{aligned}
	\end{equation}
	Compared with Eq.~\ref{eq:random-sampling}, the above DLow sampling method learns $K$ Gaussian distributions and uses the reparameterization trick to sample latent variables from these distributions: $\mathbf{z}_k \sim \mathcal{N}(\mathbf{b}_k, \mathbf{A}_k)$, and finally maps $\mathbf{z}_k$ and the input poses $\mathbf{x}$ to future poses using $\mathcal{G}_{\bm{\theta}}$ that has already been learned by the CVAE model.
	
	More formally, DLow samples a result $\mathbf{y}_k$ from the distribution of $r_{\bm{\psi}}(\mathbf{y}_k|\mathbf{x})=\int p_{\bm{\theta}}(\mathbf{y}_k|\mathbf{x},\mathbf{z}_k)r_{\bm{\psi}}(\mathbf{z}_k|\mathbf{x})d\mathbf{z}_k$ where $p_{\bm{\theta}}(\mathbf{y}_k|\mathbf{x},\mathbf{z}_k)$ is the conditional distribution modeled by $\mathcal{G}_{\bm{\theta}}$, and $r_{\bm{\psi}}(\mathbf{z}_k|\mathbf{x})$, \textit{i.e.,} $\mathcal{N}(\mathbf{b}_k,\mathbf{A}_k)$, is the latent distribution modeled by the network $\mathcal{Q}_{\bm{\psi}}$. 
	
	To train $\mathcal{Q}_{\bm{\psi}}$, DLow minimizes the following diversity loss to enlarge the distances between pairs of results predicted from the same input $\mathbf{x}$:
	\begin{equation}
		\label{eq:dlow-diversity-loss}
		\mathcal{L}_{div} = \frac{1}{K(K-1)}\sum_{i=1}^K\sum_{j\neq i}^K \exp \left( - \frac{\mathcal{D}^2(\mathbf{\tilde{y}}_i,\mathbf{\tilde{y}}_j)}{\sigma} \right),
	\end{equation}
	where $\mathbf{\tilde{y}_i}$ or $\mathbf{\tilde{y}_j}$ denotes a predicted pose sequence, and $\mathcal{D}(\cdot,\cdot)$ calculates the Euclidean distance between two predictions. Besides, the following accuracy loss is minimized:
	\begin{equation}
		\label{eq:dlow-ade-loss}
		\mathcal{L}_{acc} = \min_k \|\mathbf{y} - \mathbf{\tilde{y}}_k\|_2, k\in[1,K].
	\end{equation}
	This loss computes $\mathcal{L}_2$ distance between every prediction $\mathbf{\tilde{y}}_k$ and the ground truth $\mathbf{y}$ and returns the minimum one. Minimizing this loss makes at least one of the predictions similar to the ground truth. Finally, a Kullback-Leibler divergence loss is imposed:
	\begin{equation}
		\label{eq:kl-loss}
		\mathcal{L}_{KL} = \mathcal{KL}\left(r_{\bm{\psi}}(\mathbf{z}_k|\mathbf{x})||p(\mathbf{z})\right), k\in[1,K],
	\end{equation}
	where $p(\mathbf{z})$ is the prior distribution used to train the CVAE. This loss makes $\mathbf{z}_k$ correspond to a high-likelihood sample $\mathbf{y}_k$ under the generative model $p_{\bm{\theta}}(\mathbf{y}|\mathbf{x},\mathbf{z})$, guaranteeing the plausibility of the predicted poses.
	
	\subsection{Our method}
	\label{sec:our-method}
	While DLow improves sampling diversity compared to the random sampling method, we observe that its effectiveness is limited in two ways. (1) Firstly, DLow entangles its prediction performance with the learning of the network $\mathcal{Q}_{\bm{\psi}}$. However, the network is trained on all the training data, hence its performance is inevitably averaged over all the data, reducing its ability to make extreme predictions existing at minor modes. (2) Secondly, due to the entanglement, DLow can only sample $K$ predictions at a time. However, it is more preferable that a sampling method can sample any number of samples at test time. 
	
	To solve these problems, we propose a new sampling method which disentangles the direct correlation between the tasks of diverse prediction and the network parameter learning. Figure~\ref{fig:pipeline} illustrates the sampling process of our method. On one hand, we design a network $\mathcal{N}_{\bm{\beta}}$ parameterized by $\bm{\beta}$ that takes $\mathbf{x}$ as input and outputs a base matrix $\mathbf{B}\in \mathbb{R}^{M \times n_b}$:
	\begin{equation}
		\mathbf{B} = \mathcal{N}_{\bm{\beta}}(\mathbf{x}),
	\end{equation}
	where each row of $\mathbf{B}$ is a basis vector of dimension $n_b$ and there are $M$ basis vectors in total. One can imagine that the basis vectors together form a space which we call ``auxiliary space'' in this paper. A point in the space can be obtained by linearly combining the basis vectors. On the other hand, we use the Gumbel-Softmax sampling strategy (introduced later) to sample a coefficient matrix $\mathbf{W}\in \mathbb{R}^{K \times M}$ in which each row $\mathbf{W}_i$ ($i\in[1,K]$) contains $M$ weights used to combine the basis vectors. These weights should satisfy $\sum_{j=1}^M \mathbf{W}_{i,j} = 1$. Then, we multiply $\mathbf{W}$ and $\mathbf{B}$ together to obtain a point matrix $\mathbf{W}\mathbf{B}\in \mathbb{R}^{K \times n_b}$ where each row represents a point sampled from the auxiliary space. This operator samples $K$ points from the auxiliary space in total. Finally, we use another network $\mathcal{N}_{\bm{\gamma}}:\mathbb{R}^{K \times n_b}\rightarrow \mathbb{R}^{K \times n_z}$ parameterized by $\bm{\gamma}$ to further transform the $K$ points to $\mathbf{A_k}$ and $\mathbf{b}_k$:
	\begin{equation}
		\left\{\mathbf{A}_k,\mathbf{b}_k\right\}_{k=1}^K = \mathcal{N}_{\bm{\gamma}}(\mathbf{W}\mathbf{B}).
	\end{equation}
	
	Based on the above preparations and incorporating with the sampling process defined in Eq.~\ref{eq:dlow-sampling}, our sampling process is (Alg. \ref{alg:our-sampling-process}):
	\begin{equation}
		\label{eq:slab-sampling}
		\left\{
		\begin{aligned}
			&\bm{\epsilon}  \sim \mathcal{N}(0,1), \\
			&\mathbf{B} = \mathcal{N}_{\bm{\beta}}(\mathbf{x}), \\
			&\mathbf{W} \leftarrow \text{Gumbel-Softmax sampling}, \\
			&\left\{\mathbf{A}_k,\mathbf{b}_k\right\}_{k=1}^K = \mathcal{N}_{\bm{\gamma}}(\mathbf{W}\mathbf{B}), \\
			&\mathbf{z}_k  = \mathbf{A}_k \mathbf{\bm{\epsilon}} + \mathbf{b}_k, &1\leq k \leq K,\\
			&\mathbf{y}_k  = \mathcal{G}_{\bm{\theta}}(\mathbf{x},\mathbf{z}_k), &1\leq k \leq K.
		\end{aligned}
		\right.
	\end{equation}
	Compared with DLow which relies on a network to directly output different Gaussian distributions, our method samples the Gaussian distributions from the auxiliary space characterized by $\mathbf{B}$. At the training stage, the sampling number $K$ is fixed to train the structure of the auxiliary space and make it match with the sampling strategy (\textit{e.g.}, Gumbel-Softmax random sampling) such that the points sampled from the space by the sampling strategy can yield predictions of high diversity. At test time, since the auxiliary space and the relationship between the space and the sampling strategy has already been established, we can sample any number of points as needed. We stress that although our model is trained on all the training data, these data are used to form the shape of the auxiliary space which is flexible and adjustable to accommodate all the data.
	
	\begin{algorithm}[!t]
		\caption{Diverse sampling from a complex distribution by randomly Gumbel-Softmax sampling from an auxiliary space}\label{alg:our-sampling-process}
		\begin{algorithmic}[1]
			\Require Observed pose sequence $\mathbf{x}$, number of samples $K$, auxiliary space generation network $\mathcal{N}_{\bm{\beta}}$, Gaussian distribution generation network $\mathcal{N}_{\bm{\gamma}}$, CVAE decoder network $\mathcal{G}_{\bm{\theta}}$
			\Ensure A set of samples $\{\mathbf{\tilde{y}}_k\}_{k=1}^{K}$
			\State $\mathbf{B} = \mathcal{N}_{\bm{\beta}}(\mathbf{x})$ // \textit{generate an auxiliary space given input poses}
			\State $\mathbf{W} \leftarrow \text{Gumbel-Softmax sampling}$ // \textit{see \rm{Algorithm} \ref{alg:gumbel-softmax}}
			\State $\mathbf{P}=\mathbf{W}\mathbf{B}$ // \textit{Multiply $\mathbf{W}$ and $\mathbf{B}$ to obtain a point matrix $\mathbf{P}$}
			\State $\left\{\mathbf{A}_k,\mathbf{b}_k\right\}_{k=1}^K = \mathcal{N}_{\bm{\gamma}}(\mathbf{P})$ // \textit{convert points into means and variances}
			\State $\bm{\epsilon}  \sim \mathcal{N}(0,1)$ // \textit{sampling an $\bm{\epsilon}$ from the normal distribution}
			\For{$k=1$ to $K$}
			\State $\mathbf{z}_k  = \mathbf{A}_k \mathbf{\bm{\epsilon}} + \mathbf{b}_k$ // \textit{reparameterization trick}
			\State $\mathbf{\tilde{y}}_k  = \mathcal{G}_{\bm{\theta}}(\mathbf{x},\mathbf{z}_k)$ // \textit{decode $\mathbf{z}_k$ and $\mathbf{x}$ into a result $\mathbf{\tilde{y}}_k$}
			\EndFor 
		\end{algorithmic}
	\end{algorithm}

	In the following, we detail components of our model that help shape the auxiliary space.
	
	\subsubsection{Network Architectures}
	We have two sub-networks $N_{\bm{\beta}}$ and $N_{\bm{\gamma}}$. For $N_{\bm{\beta}}$, the input is $\mathbf{x}\in\mathbb{R}^{J\times C \times H}$ where $H$ is the length of the input sequence, $J$ is the number of joints of a pose, and each joint has $C$ coordinates. The output is $\mathbf{B}\in\mathbb{R}^{M\times n_b}$. Firstly, we use a GCN \cite{mao2019learning} to extract features in $\mathbb{R}^{J \times F}$ from $\mathbf{x}$ where $F$ is the dimension size of the features. Then, we use an MLP to map the feature map in $\mathbb{R}^{J \times F}$ to $\mathbf{B}$ in $\mathbb{R}^{M\times n_b}$. For $N_{\bm{\gamma}}$, we employ another MLP that maps a feature map in $\mathbb{R}^{K\times n_b}$ to a feature map in $\mathbb{R}^{K\times n_z}$. Please refer to the supplemental material for more details of the network designs.
	
	\subsubsection{Gumbel-Softmax Sampling}
	We randomly sample a coefficient matrix $\mathbf{W}$ by the Gumbel-Softmax sampling method (Alg. \ref{alg:gumbel-softmax}) by which each row $\mathbf{W}_i$ ($i\in [1,K]$) of $\mathbf{W}$ is calculated as:
	\begin{equation}
		\left\{
		\begin{aligned}
			&u_{ij} \sim \mathrm{U}(0, 1), j \in [1, M],\\
			&g_{ij} = -\mathrm{log}(-\mathrm{log}(u_{ij})), j \in [1, M],\\
			&\mathbf{W}_{ij} = \frac{\pi + g_{ij}}{\tau}, j \in [1, M],\\
			&\mathbf{W}_i = \mathrm{Softmax}(\mathbf{W}_i)
		\end{aligned}
		\right. 
		\label{eq:gumbel}
	\end{equation} 
	where $\mathrm{U}(0, 1)$ is the uniform distribution, $\pi$ and $\tau$ are parameters of the Gumbel distribution which are set to $1/M$ and 1, respectively. 

	Besides the Gumbel distribution, we can also sample from a uniform or Gaussian distribution at first and then apply the Softmax normalization to obtain a coefficient matrix. However, Gumbel-Softmax sampling is more aggressive than Uniform-Softmax and Gaussian-Softmax sampling in assigning larger weights for a few basis vectors while making other basis vectors sharing just a small portion of the weight. In other words, the Gumbel-Softmax sampling strategy can samples points more near to the basis vectors. This benefits the learning of the basis vectors, because the network only needs to diversify the basis vectors to obtain diverse Gaussian distributions and eventually generate diverse poses. Please see our ablation study of comparisons among them.
	

	\begin{algorithm}[!t]
		\caption{Gumbel-Softmax coefficient matrix generation}\label{alg:gumbel-softmax}
		\begin{algorithmic}[1]
			\Require Number of coefficient vectors $K$, dimension size $M$ of a coefficient vector, Gumbel distribution parameters $\pi$ and $\tau$
			\Ensure A coefficient matrix $\mathbf{W} \in \mathbb{R}^{K\times M}$
			\State Declare a matrix $\mathbf{W} \in \mathbb{R}^{K\times M}$
			\For{$i=1$ to $K$}
			\For{$j=1$ to $M$}
			\State $u \sim \mathrm{U}(0, 1)$ // \textit{sample a value from uniform distribution}
			\State $g = -\mathrm{log}(-\mathrm{log}(u))$
			\State $\mathbf{W}_{ij} = \frac{\pi + g}{\tau}$
			\EndFor
			\State $\mathbf{W}_i = \mathrm{Softmax}(\mathbf{W}_i)$ // \textit{normalize the $i^{th}$ row of $\mathbf{W}$}
			\EndFor
		\end{algorithmic}
	\end{algorithm}

	\begin{table*}[!t]
		\caption{Quantitative comparisons. All the results are calculated by sampling 50 times for each input historical pose sequence. The best results are marked in bold.}
		\label{tab:main_result}
		\resizebox{1\textwidth}{!}{
			\begin{tabular}{c|c|ccccc|ccccc}
				\toprule
				& \multirow{2}{*}{Method}   & \multicolumn{5}{c|}{Human3.6M \cite{ionescu2013human3}}         & \multicolumn{5}{c}{HumanEva-I \cite{sigal2010humaneva}}        \\ \cline{3-12} 
				&                           & APD $\uparrow$   & ADE $\downarrow$  & FDE $\downarrow$  & MMADE $\downarrow$ & MMFDE $\downarrow$ & APD  $\uparrow$ & ADE $\downarrow$  & FDE  $\downarrow$ & MMADE $\downarrow$ & MMFDE $\downarrow$ \\ \hline
				\multirow{2}{*}{deterministic} & LTD \cite{mao2019learning}  & 0.000  & 0.516 & 0.756 & 0.627 & 0.795 & 0.000 & 0.415 & 0.555 & 0.509 & 0.613 \\
				& MSR \cite{dang2021msr}                  & 0.000  & 0.508 & 0.742 & 0.621 & 0.791 & 0.000 & 0.371 & 0.493 & 0.472 & 0.548 \\ \hline
				\multirow{12}{*}{stochastic}   & Pose-Knows \cite{walker2017pose}              & 6.723  & 0.461 & 0.560 & 0.522 & 0.569 & 2.308 & 0.269 & 0.296 & 0.384 & 0.375 \\
				&MT-VAE \cite{yan2018mt}                  & 0.403  & 0.457 & 0.595 & 0.716 & 0.883 & 0.021 & 0.345 & 0.403 & 0.518 & 0.577 \\
				&HP-GAN \cite{barsoum2018hp}                  & 7.214  & 0.858 & 0.867 & 0.847 & 0.858 & 1.139 & 0.772 & 0.749 & 0.776 & 0.769 \\ 
				&BoM \cite{bhattacharyya2018accurate}                     & 6.265  & 0.448 & 0.533 & 0.514 & 0.544 & 2.846 & 0.271 & 0.279 & 0.373 & 0.351 \\
				&GMVAE \cite{dilokthanakul2016deep}                   & 6.769  & 0.461 & 0.555 & 0.524 & 0.566 & 2.443 & 0.305 & 0.345 & 0.408 & 0.410 \\
				&DeLiGAN \cite{gurumurthy2017deligan}                 & 6.509  & 0.483 & 0.534 & 0.520 & 0.545 & 2.177 & 0.306 & 0.322 & 0.385 & 0.371 \\
				&DSF \cite{yuan2019diverse}                     & 9.330  & 0.493 & 0.592 & 0.550 & 0.599 & 4.538 & 0.273 & 0.290 & 0.364 & 0.340 \\
				&DLow \cite{yuan2020dlow}                   & 11.741 & 0.425 & 0.518 & 0.495 & 0.531 & 4.855 & 0.251 & 0.268 & 0.362 & 0.339 \\
				&GSPS \cite{mao2021generating}                    & 14.757 & 0.389 & {0.496} & 0.476 & {0.525} & 5.825 & 0.233 & 0.244 & 0.343 & 0.331 \\ \cline{2-12} 
				& Ours             &    \textbf{15.310}   &  \textbf{0.370}    & \textbf{0.485}  &    \textbf{0.475}   &  \textbf{0.516}    &   \textbf{6.109}   &  \textbf{0.220}     &    \textbf{0.234}   &   \textbf{0.342}    &    \textbf{0.316}   \\ 
				\bottomrule
		\end{tabular}}
	\end{table*}

	\begin{figure*}[!t]
		\centering
		\includegraphics[width=1.0\linewidth]{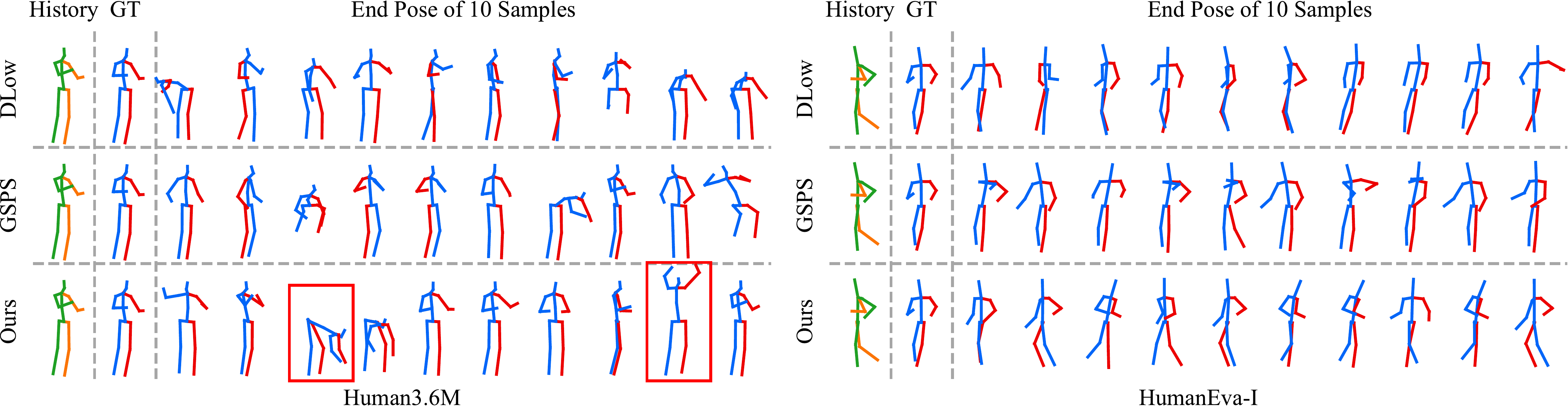}
		\caption{Qualitative comparisons. For the same input, we show end poses of 10 predicted results. Please see actions of the poses.}
		\label{fig:quality_visualize}
	\end{figure*}
	
	\subsubsection{Training Losses} 
	Let $\{\mathbf{\tilde{y}}_k\}_{k=1}^K$ be the $K$ results predicted from an input, we impose three kinds of loss functions on $\{\mathbf{\tilde{y}}_k\}_{k=1}^K$ to train the proposed sampling framework.
	
	(1) \textit{Hinge-diversity loss.} In order to enhance the diversity of the results, we propose the following hinge-diversity loss:
	\begin{equation}
		\mathcal{L}_{hdiv} = \frac{1}{K(K-1)} \sum_{i=1}^{K} \sum_{j\neq i}^{K} \max \left( 0, \eta -  \left\| \mathbf{\tilde{y}}_i - \mathbf{\tilde{y}}_j \right\|_2 \right),
		\label{equ:hinge_loss}
	\end{equation}
	where $\eta$ is a user-defined threshold. By the hinge-diversity loss, we explicitly enforce the distance between any pair of generated predictions to be no less than $\eta$. Compared with the diversity loss defined in Eq.~\ref{eq:dlow-diversity-loss}, the hinge-diversity loss is more aggressive in enforcing the diversity of the predictions while less affecting the accuracy of the results (see ablation studies).
	
	(2) \textit{Accuracy loss.} To ensure the accuracy of results, we also adopt the accuracy loss $\mathcal{L}_{acc}$ defined in Eq.~\ref{eq:dlow-ade-loss} that enforces at least one of the predictions to be similar to the ground truth. 
	
	(3) \textit{KL loss.} Finally the KL loss defined in Eq.~\ref{eq:kl-loss} is a very important loss which ensures the model to produce realistic and plausible results instead of those with high diversity but are physically invalid. Our KL loss is now defined as:
	\begin{equation}
		\mathcal{L}^\prime_{KL} = \mathcal{KL} \left(r_{\bm{\beta},\bm{\gamma}}(\mathbf{z}_k|\mathbf{x})||p(\mathbf{z})\right), k\in[1,K],
	\end{equation}
	where $r_{\bm{\beta},\bm{\gamma}}(\mathbf{z}_k|\mathbf{x})$ is the latent distribution of $\mathbf{z}_k$ encoded by networks $N_{\bm{\beta}}$ and $N_{\bm{\gamma}}$ with parameters of $\bm{\beta}$ and $\bm{\gamma}$. 
	
	Altogether, our training loss is:
	\begin{equation}
		\begin{aligned}
			\mathcal{L} &= \lambda_{hdiv} \mathcal{L}_{hdiv} + \lambda_{acc} \mathcal{L}_{acc} + \lambda_{KL} \mathcal{L}^\prime_{KL},
		\end{aligned}
		\label{equ:all_loss_one_stream}
	\end{equation}
	where $\lambda$s are hyper-parameters used to balance the three terms.
	
	\section{Experiments}
	
	
	\subsection{Experimental Settings}
	
	\begin{figure*}[!t]
		\centering
		\includegraphics[width=1\linewidth]{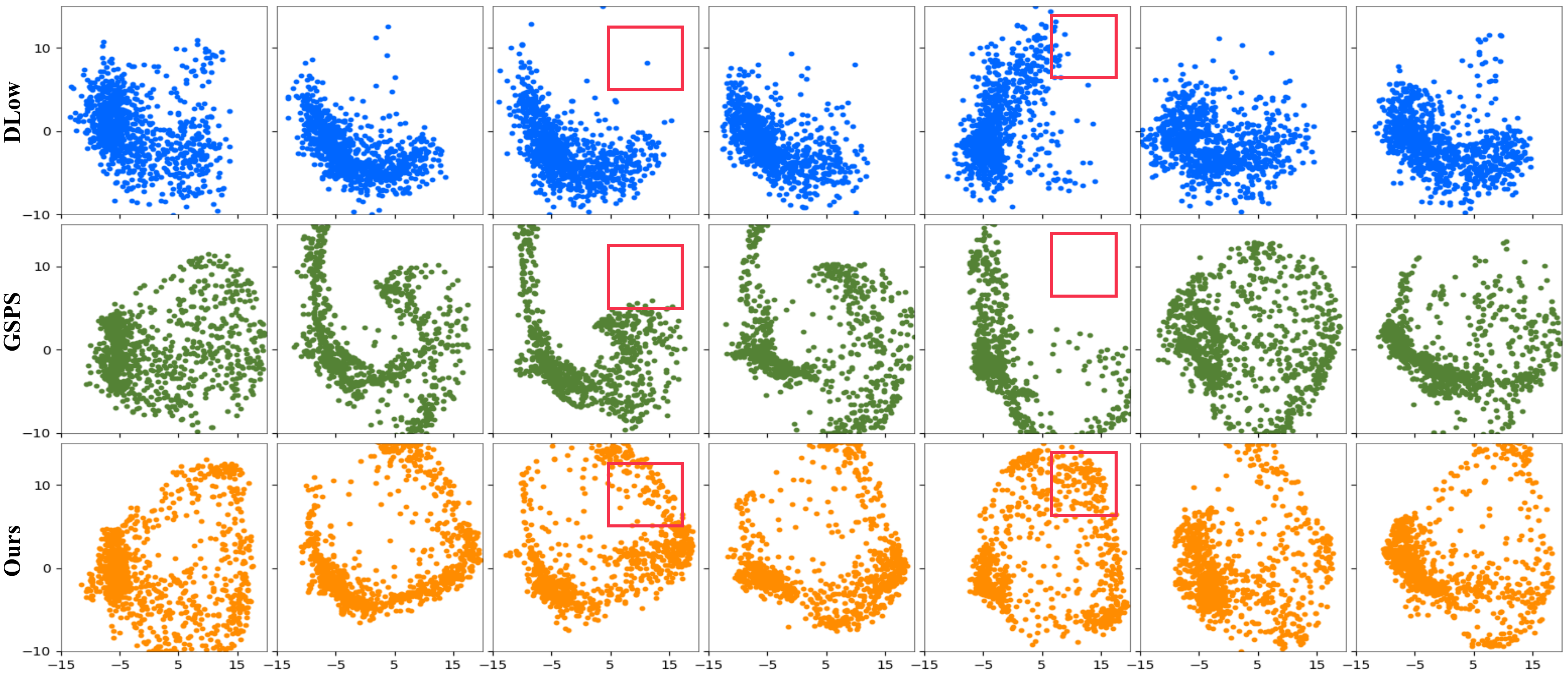}
		\caption{Holistic views of results. 1000 pose sequences are predicted and projected to 2D points. Note the regions marked with red boxes where our method can sample points from while DLow and GSPS fail to.}
		\label{fig:pca_diversity}
	\end{figure*}
	
	\textbf{Datasets.} Following \cite{yuan2020dlow, mao2021generating}, we evaluate our method on two public motion capture datasets: Human3.6M\footnote{The authors Lingwei Dang and Yongwei Nie signed the license and produced all the experimental results in this paper. Meta did not have access to the Human3.6M dataset.}~\cite{ionescu2013human3} and HumanEva-I\footnote{We strictly comply with the agreement of using this dataset for non-commercial research purpose only.}~\cite{sigal2010humaneva}. (1) \textbf{Human3.6M} contains 7 subjects each performing 15 action categories. We use the data of five subjects (S1, S5, S6, S7, S8) for training, and the other two (S9, S11) for testing. After removing redundant joints, each pose has 17 joints. We input 25 frames, \textit{i.e.}, 0.5s (50fps), to forecast 100 frames (2s) in the future. (2) \textbf{HumanEva-I} comprises 3 subjects each performing 5 action categories. Each pose has 15 joints. We forecast 60 future poses (1s, 60fps) given 15 (0.25s) frames. 
	
	\textbf{Evaluation Metrics.} We use five metrics to evaluate our method. (1) \textbf{APD}: the Average Pairwise Distance of results predicted from an input \cite{aliakbarian2020stochastic}. This metric measures the diversity of the results. (2) \textbf{ADE} and \textbf{FDE}: ADE computes the Average Displacement Error between the ground truth and the result most similar to the ground truth. FDE, which stands for Final Displacement Error, only calculates the distance between the last pose of GT and the last pose of the most similar result to GT. (3) \textbf{MMADE} and \textbf{MMFDE} are the multimodal versions of ADE and FDE which were introduced in \cite{yuan2020dlow}. To compute them, for each training sample $(\mathbf{x},\mathbf{y})$, one needs to search the whole dataset for a set of $\{(\mathbf{x}_p,\mathbf{y}_p)\}_{p=1}^P$ whose past motion $\mathbf{x}_p$ is similar enough to $\mathbf{x}$, and take their future motion $\{\mathbf{y}_p\}_{p=1}^{P}$ as the pseudo ground truths of $\mathbf{x}$. MMADE is then computed as: $\frac{1}{P} \sum_{p=1}^{P} \min_{i} \left\|  \mathbf{\tilde{y}}_i - \mathbf{y}_p \right\|_2, i \in \{1, \cdots, K\}$. 
	Similar to FDE, MMFDE only calculates the error of end poses. By ADE, FDE, MMADE, and MMFDE, we can know the accuracy of results.
	
	We employ additional metrics, \textit{e.g.}, ADE-m, FDE-m, ACC, FID suggested by \cite{bie2022hit} for further evaluation. please refer to the supplementary material for more details.
	

	\textbf{Implementation Details.} 
	In default, we use $M=40$ basis vectors. At training time, we sample $K=50$ points from the auxiliary space. At test time, we set $K=50$ to compare with previous approaches, and set $K$ to numbers from 2 to 1000 in ablation studies. We set $n_b = 128$, and $n_z = 64$. For Human3.6M, we set $\lambda_{hdiv} =20$, $\lambda_{acc}=40$, and $\lambda_{KL}=0.5$, and $\eta$ in Eq.~\ref{equ:hinge_loss} to 25. These numbers for the HumanEva-I dataset are 100, 25, 0.1 and 20, respectively. 
	
	We implement our method in PyTorch, training it by the Adam optimizer with a learning rate of $1e-3$ for the first 100 training epochs. Then the learning rate starts to decrease, eventually becoming $7e-4$ after a total of 500 epochs of training. Following \cite{yuan2020dlow,mao2021generating}, for each epoch, we randomly sample 5000 samples from Human3.6M or 2000 samples from HumanEva-I for training. The batchsize is set to 16 for both datasets.
	
	\begin{table*}[!t]
		\caption{We perform 5 groups of ablation studies. $*$ indicates default choices. Please refer to the main text for details.}
		\label{tab:ablation}
		\resizebox{1\textwidth}{!}{
			\begin{tabular}{c|ccccc|ccccc|ccc|cc|ccc}
				\toprule
				& \multicolumn{5}{c|}{ $\textcircled{1}$ Number of basis vectors}                                & \multicolumn{5}{c|}{$\textcircled{2}$ Dimension of auxiliary space} & \multicolumn{3}{c|}{$\textcircled{3}$ Sampling method}                    & \multicolumn{2}{c|}{$\textcircled{4}$ $N_{\bm{\gamma}}$}         & \multicolumn{3}{c}{$\textcircled{5}$ $\mathcal{L}_{hdiv}$ \textit{v.s.} $\mathcal{L}_{div}$} \\ \cline{2-19}
				& 20    & 30     & $40^*$ & 50     & \multicolumn{1}{c|}{60}     & 32 & 64 & ${128}^*$ & 256 & \multicolumn{1}{c|}{512} & ${\mathrm{Gumbel}}^*$  & Gaussian & \multicolumn{1}{c|}{Uniform} &  ${\mathrm{w/} N_{\bm{\gamma}}}^*$   & \multicolumn{1}{c|}{w/o $N_{\bm{\gamma}}$} & ${\mathcal{L}_{hdiv}(25)}^*$  & $\mathcal{L}_{div}$ (25) & $\mathcal{L}_{div}$ (1300)  \\ \hline
				APD $\uparrow$  & 5.929 & 5.946 & \textbf{5.993}   & 5.969 & \multicolumn{1}{c|}{5.951} & 5.885 & 5.931 & \textbf{5.993} & 5.963 & \multicolumn{1}{c|}{5.957} & \textbf{5.993} & 5.847    & \multicolumn{1}{c|}{5.730}   & \textbf{5.993}& \multicolumn{1}{c|}{5.182}  & \textbf{5.993}     & 3.843         & \textbf{5.993}  \\
				ADE $\downarrow$ & 0.234 & 0.234 & 0.231   & \textbf{0.229} & \multicolumn{1}{c|}{0.233} & 0.236 & 0.233 & \textbf{0.231} & 0.233 &  \multicolumn{1}{c|}{0.236} & \textbf{0.231} & 0.240    & \multicolumn{1}{c|}{0.233}   & 0.231 & \multicolumn{1}{c|}{\textbf{0.229}}  & 0.231     & \textbf{0.211}        & 0.235  \\
				FDE $\downarrow$ & 0.240 & 0.243 & 0.240   & \textbf{0.239}& \multicolumn{1}{c|}{0.241} & 0.245 & 0.243 & \textbf{0.240} & 0.241 & \multicolumn{1}{c|}{0.243} & \textbf{0.240}& 0.246    & \multicolumn{1}{c|}{0.241}   & 0.240 & \multicolumn{1}{c|}{\textbf{0.236}}  & 0.240     & \textbf{0.218}         & 0.242  \\
				MMADE $\downarrow$ & 0.345 & 0.345 & 0.340   & 0.339 & \multicolumn{1}{c|}{\textbf{0.338}} & 0.337 & 0.342 & 0.340 & \textbf{0.336} & \multicolumn{1}{c|}{0.342} & \textbf{0.340} & 0.343    & \multicolumn{1}{c|}{0.344}   & 0.340 & \multicolumn{1}{c|}{\textbf{0.322}}  & 0.340     & \textbf{0.309}        &0.343  \\
				MMFDE $\downarrow$  & 0.321 & 0.319 & 0.313   & 0.315 & \multicolumn{1}{c|}{\textbf{0.312}} & 0.312 & 0.318 & 0.313 & \textbf{0.310} & \multicolumn{1}{c|}{0.316} & \textbf{0.313} & 0.320    & \multicolumn{1}{c|}{0.323}   & 0.313 & \multicolumn{1}{c|}{\textbf{0.298}}  & 0.313     & \textbf{0.287}           & 0.321   \\ 
				\bottomrule
		\end{tabular}}
	\end{table*}

	\begin{figure*}[!t]
		\centering
		\includegraphics[width=0.9\linewidth]{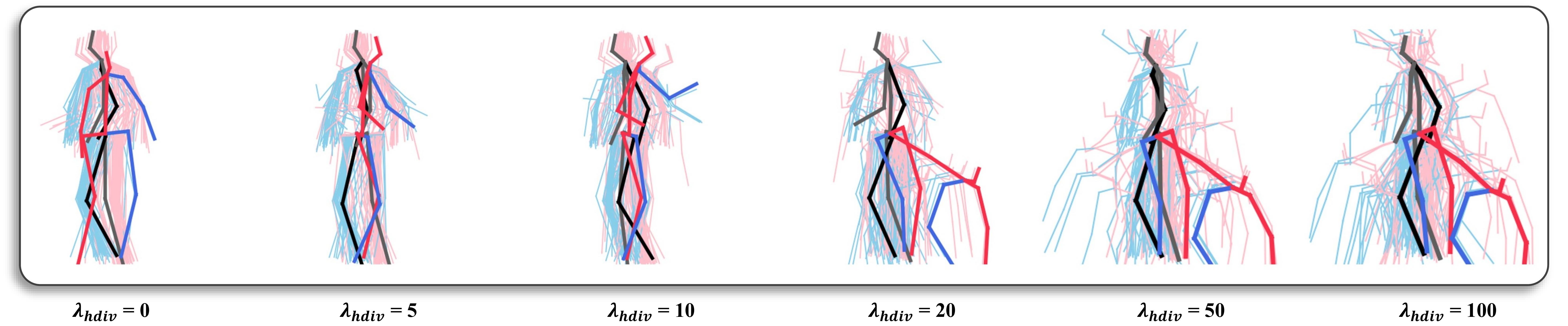}
		\caption{Increase $\lambda_{hdiv}$ from 0 to 100. The last poses of 50 pose sequences predicted from an input are stacked together to illustrate the holistic view of results. Poses most similar to and different from the ground truth are highlighted.}
		\label{fig:qian_shou_guan_yin}
	\end{figure*}
	
	\subsection{Comparison with Previous Approaches}
	
	We compare our method with both kinds of prediction methods: 
	(1) The most recent deterministic methods including LTD \cite{mao2019learning} and MSR \cite{dang2021msr}. 
	(2) Stochastic methods including HP-GAN \cite{barsoum2018hp}, Pose-Knows \cite{walker2017pose}, MT-VAE \cite{yan2018mt}, BoM \cite{bhattacharyya2018accurate}, GMVAE \cite{dilokthanakul2016deep}, DeLiGAN \cite{gurumurthy2017deligan}, DSF \cite{yuan2019diverse}, DLow \cite{yuan2020dlow}, and GSPS \cite{mao2021generating}. For each input historical pose sequence, all stochastic methods predict 50 different future sequences.
	
	Table \ref{tab:main_result} shows comparisons among all the compared methods. On both datasets, our method outperforms all the other approaches on all the evaluation metrics. Since deterministic approaches can only generate one output, the diversity of their results is 0.000. In terms of prediction accuracy, deterministic methods are inferior to stochastic methods too. This may be due to two reasons. Firstly, deterministic approaches are not good at long-term prediction (\textit{e.g.}, more than 1 second). Secondly, stochastic methods can predict multiple results among which there may be a very good one. 
	Since our method is based on DLow, let us focus on the comparisons between them. On human3.6M, DLow achieves a diversity of 11.741, while that of our method is 15.310, which is a very significant improvement of about 30\%. Our method is also better than DLow in terms of prediction accuracy. For ADE, our accuracy is improved by 14.9\% ( 0.370 \textit{v.s.} 0.425). For FDE, MMADE, MMFDE, the improvements are: 6.8\%, 4.2\%, and 2.9\%, respectively. The comparisons on the HumanEva-I dataset show similar trends: our method improves DLow on all metrics. GSPS is one of the latest stochastic prediction methods which directly predicts very diverse results as long as they are reasonable under many prior constraints. Our method outperforms GSPS, reaching a new state-of-the-art.
	
	
	To show the quality of the predicted poses, we visualize end poses of pose sequences predicted by DLow \cite{yuan2020dlow}, GSPS \cite{mao2021generating} and our method. The examples in Figure \ref{fig:quality_visualize} show that our method produces more diverse results than DLow and GSPS. For example, on the left, our method can predict actions of ``raising hands'' and ``picking up things'' (marked by red boxes). Please refer to the supplemental video for how our method smoothly transitions the action from the input ``normal standing'' to the two very different actions.
	
	Figure \ref{fig:pca_diversity} illustrates the holistic views of results. Given an input, we generate 1000 results and project them into 2D space. Note that DLow can only sample 50 results at a time. To generate 1000 results for DLow, we repeatedly run DLow 20 times. 
	As can be seen, our results occupy the 2D space more evenly. Please compare regions marked by red boxes where our method can sample points from while DLow and GSPS cannot.

	\subsection{Ablation Study}
	\label{sec:ablation}
	
	
	Table~\ref{tab:ablation} shows five groups of ablation studies that validate the design components of our method. All models are trained on the HumanEva-I dataset for 200 epochs.
	
	(1) \textbf{Number of basis vectors.} We set $M$ to 20, 30, 40, 50, and 60. Overall, $M$ has little effect on the results. For example, APD (\textit{i.e.}, diversity) varies within a narrow range of [5.929, 5.993], with the best diversity obtained when $M=40$. MMADE and MMFDE steadily improve as $M$ increases. We finally choose 40 as the default value of $M$.
	(2) \textbf{Dimension size of auxiliary space.} We set $n_b$ to 32, 64, 128, 256, and 512. The best APD, ADE, and FDE are obtained when $n_b=128$. For MMADE and MMFDE, the best values are obtained when $n_b=256$. We finally choose 128 as the default value of $n_b$.
	(3) \textbf{Gumbel \textit{v.s.} Gaussian and Uniform.} The experimental results validate that Gumbel-Softmax sampling strategy is better than Uniform-Softmax and Gaussian-Softmax sampling methods when used in our sampling process. The fact that the differences between different sampling methods is not evident can be ascribed to the high capability of the proposed auxiliary-space-based resampling method. The space itself can be flexibly adjusted to match the three sampling methods to output good results. 
	(4) \textbf{Using $\mathcal{N}_{\bm{\gamma}}$ or not.} In this ablation study, we remove the second MLP network $\mathcal{N}_{\bm{\gamma}}$, and directly use each row of the point matrix as a pair of $\mathbf{A}_k$ and $\mathbf{b}_k$. The diversity drops significantly (expected but undesired), while the accuracy increases. This is reasonable as diversity and accuracy are two conflict objectives: the increase of the diversity inevitably decreases the accuracy. We ultimately choose to integrate the network into our framework to achieve higher diversity with acceptable accuracy.
	(5) \textbf{Diversity loss \textit{v.s.} hinge-diversity loss.} We replace our hinge-diversity loss with the diversity loss defined in Eq.~\ref{eq:dlow-diversity-loss}. Note that the default weight of our hinge-diversity loss is $\lambda_{hdiv}=25$. For the diversity loss, we first use a weight of 25. However, the diversity is much lower: 3.843 \textit{v.s.} our 5.993. We increase the weight of the diversity loss to 1300 until it produces the same diversity as ours, but now its accuracy is lower than that of our hinge-diversity loss. These studies show our hinge-diversity loss better prompts diversity while less affecting the accuracy.

	\begin{figure}[!t]
		\centering
		\includegraphics[width=0.485\linewidth]{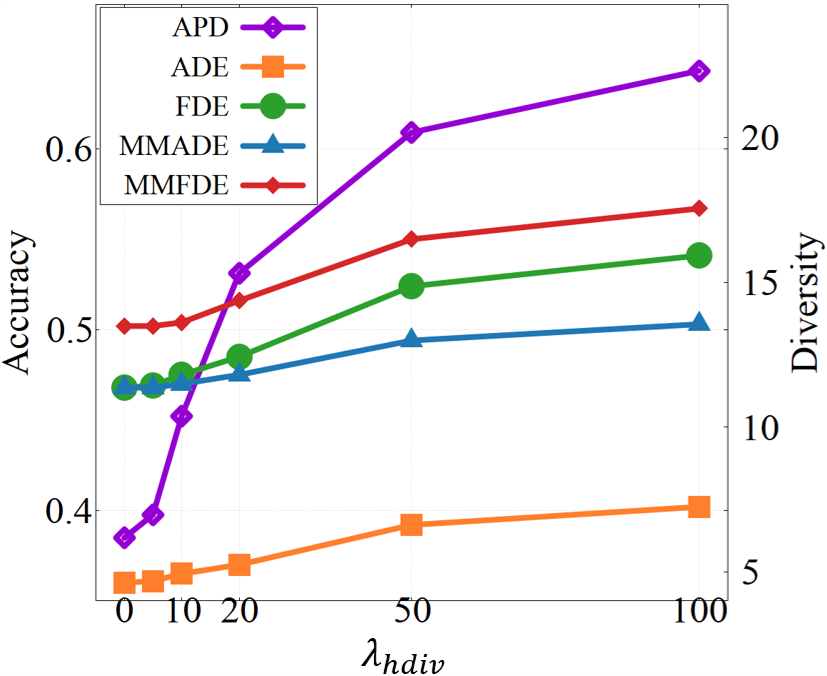}
		\hspace{0.1cm}
		\includegraphics[width=0.485\linewidth]{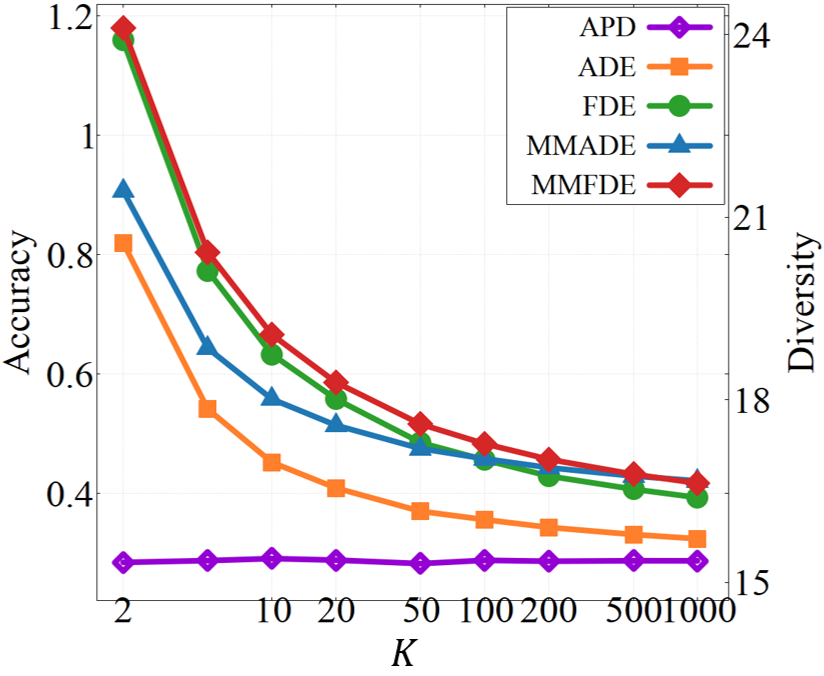} \\
		\flushleft \hspace{2cm} (a) \hspace{3.5cm} (b)
		\caption{(a) As $\lambda_{hdiv}$ increases, APD (diversity) drastically increases from 6.174 to 22.300, while accuracy decreases. (b) As $K$ increases, accuracy becomes better while diversity nearly does not change.}
		\label{fig:two_curves}
	\end{figure}
	
	In Figure \ref{fig:qian_shou_guan_yin}, we perform an ablation study on the weight of the hinge-diversity loss. We set $\lambda_{hdiv}$ to 0, 5, 10, 20, 50, and 100. All models are trained on Human3.6M for 200 epochs. We stack the end poses of all the results predicted from an input. The pose most similar to (in black and gray) and different from (red and blue) the ground truth are highlighted. Visually, as $\lambda_{hdiv}$ increases, more diverse results are obtained. However, Figure~\ref{fig:two_curves} (a) shows that larger $\lambda_{hdiv}$ leads to lower accuracy. We finally choose 20 as the default value of $\lambda_{hdiv}$, obtaining both satisfactory diversity and accuracy.
	
	In Figure~\ref{fig:two_curves} (b), we increase $K$ from 2 to 1000. As $K$ increases, ADE, FDE, MMADE and MMFDE all decrease, meaning more accurate results are obtained. This indicates that we can obtain more accurate results by sampling more of them. The diversity nearly does not change as $K$ increases, which is a good property as we can obtain diverse results with just a few samplings.
	
	
	\subsection{Limitations and Future Work}
	One limitation is that we have to adjust the weights of the loss functions to make a tradeoff between diversity and accuracy, though we note that DLow and GSPS suffer from this limitation too. Another limitation is that similar to DLow and GSPS our method occasionally generates odd poses with such as slightly long bones or unnatural actions. In the future, we can add more regularization terms, such as the bone and angle constraints adopted by GSPS, into our model to prevent these failures.
	
	\section{Conclusion}
	We have presented a diverse pose prediction algorithm. Our method first generates an auxiliary space from the input, then samples points from the auxiliary space by the Gumbel-Softmax sampling strategy, and finally maps the points to Gaussian distributions from which we sample latent codes and finally decode them into target predictions. We have demonstrated the influence of the dimension size of the auxiliary space and the number of basis vectors that characterize the auxiliary space on the performance of our method. We have also illustrated the effectiveness of the proposed hinge-diversity loss in promoting diversity while persisting accuracy. Although we only apply this method to tackle the task of stochastic human motion prediction, we believe it can also be used to handle many other stochastic prediction/generation problems.
	
	\begin{acks}
		This work is sponsored by Prof. Yongwei Nie's and Prof. Guiqing Li's Natural Science Foundation of China projects (62072191, 61972160), and their Natural Science Foundation of Guangdong Province projects (2019A1515010860, 2021A1515012301).
	\end{acks}
	
	\bibliographystyle{ACM-Reference-Format}
	\bibliography{diverse_sampling_arxiv}
	
	\clearpage
	
	\appendix
	
	\noindent{\LARGE{\textbf{Supplementary Material}}}

	
	In this supplementary material, we provide more information that cannot be included in the paper due to the space limit. We first introduce network architectures of our method in detail. Then, we provide more quantitative and qualitative comparisons. Finally, we give some failure cases and the reasons for these failures. Please refer to our provided video demo to review the results more intuitively.

	\section{Auxiliary-space-based Sampling Network Architecture}

\begin{figure*}[!t]
		\centering
		\includegraphics[width=\textwidth]{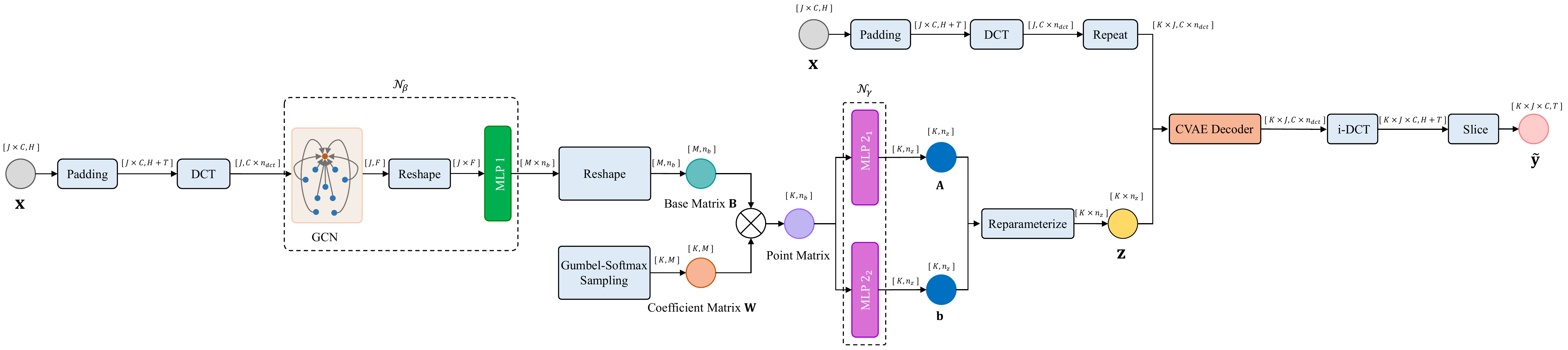}\\
		\caption{Detailed architecture of our auxiliary-space-based sampling model. A circle represents input, intermediate, or output data. The symbol above a circle indicates the size of the data. For example, [$J\times C$, $H$] means the data is a two-dimensional matrix of $J\times C$ rows and $H$ columns. The symbol above an arrow indicates the size of the output of the corresponding previous operator. Please refer to the main text for detailed descriptions of the architecture. }
		\label{fig:our-arch}
	\end{figure*}

	Our auxiliary-space-based sampling network is illustrated in Figure \ref{fig:our-arch}. The input is $\mathbf{x}\in\mathbb{R}^{[J\times C, H]}$, where $H$ is the number of input poses, $J$ is the number of joints of a pose, and $C$ is the dimension size of each joint. For Human3.6M \cite{ionescu2013human3}, $J=17$, $C=3$, $H=25$, while for HumanEva-I \cite{sigal2010humaneva}, $C=3$, $J=15$, and $H=15$.
	Following \cite{mao2019learning}, we repeat the last pose of $\mathbf{x}$, $T$ times, and append them to $\mathbf{x}$. Now, the input is of size $[J\times C, H+T]$, where $T$ is the number of poses to be predicted. 
	For Human3.6M, $T = 100$, and for HumanEva-I, $T = 60$.
	Then following \cite{mao2019learning} again, we apply a Discrete Cosine Transform (DCT) operator to
	transform the temporal information along the $H+T$ dimension of the input data into the frequency space. By keeping only the coefficients of low frequency components and discarding those of high frequency components, we obtain data of $[J\times C, n_{dct}]$ which becomes $[J, C\times n_{dct}]$ after reshaping, where $n_{dct} = 10$ is the number of the remained coefficients.
	Following, a network $\mathbf{\mathcal{N}_{\beta}}$ made up of a GCN and an MLP learns a base matrix $\mathbf{B} \in \mathbb{R}^{[M \times n_b]}$ from the DCT coefficients, where $M = 40$ and $n_b=128$.
	The GCN, which will be described later, extracts hidden features of shape $J \times F$ where $F = 256$ is the feature dimension size. Then, the MLP composed of a linear transformation layer, a Batch Normalization (BN) layer and a Tanh activation function, maps the hidden features into $\mathbf{B}$. The network structures of the GCN and MLP are shown in Table \ref{tab:component_ours}.
	Next, we sample a random coefficient matrix $\mathbf{W} \in \mathbb{R}^{[K \times M]}$ by the Gumbel-Softmax sampling technique, where $K=50$ is the sampling number. Then the multiplication of $\mathbf{W}$ and $\mathbf{B}$ outputs a point matrix of shape $[K \times n_b]$.
	Next, the second network $\mathcal{N}_{\gamma}$ projects the $K$ sampled points into the parameters of $K$ Gaussian distributions $\{\mathcal{N}( \mathbf{b}_k, \mathbf{A}_k )\}_{k=1}^{K}$, where $\mathbf{b} \in \mathbb{R}^{[K \times n_z]}$ indicates means of these Gaussian distributions and $\mathbf{A} \in \mathbb{R}^{[K \times n_z]}$ are the diagonal values of their co-variance matrices.
	In particular, $\mathcal{N}_{\gamma}$ consists of two sub MLPs for generating $\mathbf{b}$ and $\mathbf{A}$, respectively. The detailed structure of $\mathcal{N}_{\gamma}$ is shown in Table \ref{tab:component_ours}. Each MLP is made up of a Linear-BN-Tanh layer to transform the point features into hidden vectors of shape $[K \times n_h]$, and another Linear layer that maps the hidden vectors into Gaussian parameters of shape $[K \times n_z]$, where $n_h$ and $n_z$ are both set as 64.
	After that, a set of latent variables $\mathbf{z} \in \mathbb{R}^{[K \times n_z]}$ are drawn from the Gaussian distributions by the reparameterization trick.
	We repeat the input data after DCT, $K$ times and concatenate each of them with $\mathbf{z}$, and feed them into the pretrained CVAE decoder (which will be described later) to produce future motions in the frequency space of shape $[K \times J, C \times n_{dct}]$. Then we project the frequency features back into the pose space by the inverse DCT (i-DCT) function, obtaining $K$ pose sequences of shape $[K \times J \times C, H+T]$.
	Finally, a slice operator extracts only the future $T$ frames and outputs results of shape $[K \times J \times C, T]$ which are the $K$ future pose sequences predicted by our network.
	
	\begin{table}[!t]
		\caption{The network structure of $\mathbf{\mathcal{N}_{\beta}}$ and $\mathbf{\mathcal{N}_{\gamma}}$. For Human3.6M, $J=17$. For HumanEva-I, $J=15$. For both datasets, $M=40$, $K=50$, $C=3$, $F=256$, $n_{dct}=10$, $n_b=128$, $n_h=64$, $n_z=64$.}
		\label{tab:component_ours}
		\resizebox{0.45\textwidth}{!}{
			\begin{tabular}{c|ccc|c|c}
				\toprule
				Component                    & \multicolumn{1}{c|}{Block}                         & \multicolumn{1}{c|}{Layer}     & Weight Size                 & Input Size             & Output Size              \\ 
				\midrule
				\multirow{12}{*}{$\mathcal{N}_{\beta}$}          & \multicolumn{1}{c|}{\multirow{10}{*}{GCN}}         & \multicolumn{1}{c|}{GCL}     & A($J, J$), W($C \times n_{dct}, F$)  & ($J, C \times n_{dct}$)    & ($J, F$)                 \\ \cline{3-6} 
				& \multicolumn{1}{c|}{}                              & \multicolumn{1}{c|}{BN, Tanh} &  -                    & ($J, F$)   & ($J, F$)                 \\ \cline{3-6} 
				& \multicolumn{1}{c|}{}                              & \multicolumn{1}{c|}{GCL}      & A($J, J$), W($F, F$) & ($J, F$)   & ($J, F$)                 \\ \cline{3-6} 
				& \multicolumn{1}{c|}{}                              & \multicolumn{1}{c|}{BN, Tanh} &  -                     & ($J, F$)   & ($J, F$)                 \\ \cline{3-6}  
				& \multicolumn{1}{c|}{}                              & \multicolumn{1}{c|}{GCL}      & A($J, J$), W($F, F$) & ($J, F$)   & ($J, F$)                 \\ \cline{3-6} 
				& \multicolumn{1}{c|}{}                              & \multicolumn{1}{c|}{BN, Tanh} &  -                      & ($J, F$)   & ($J, F$)                 \\ \cline{3-6} 
				& \multicolumn{1}{c|}{}                              & \multicolumn{1}{c|}{GCL}     & A($J , J$), W($F, F$) & ($J, F$)   & ($J, F$)                 \\ \cline{3-6}
				& \multicolumn{1}{c|}{}                              & \multicolumn{1}{c|}{BN, Tanh}  & -                       & ($J, F$)   & ($J, F$)                 \\ \cline{3-6} 
				& \multicolumn{1}{c|}{}                              & \multicolumn{1}{c|}{GCL}      & A($J , J$), W($F, F$) & ($J, F$)   & ($J, F$)                 \\ \cline{3-6} 
				& \multicolumn{1}{c|}{}                              & \multicolumn{1}{c|}{BN, Tanh}  & -                       & ($J, F$)   & ($J, F$)                 \\ \cline{2-6} 
				
				& \multicolumn{1}{c|}{\multirow{2}{*}{MLP 1}}  & \multicolumn{1}{c|}{Linear}  & W( $J \times F$, $M \times n_b$)       & ($J \times F$)    & ($M \times n_b$)                 \\ \cline{3-6} 
				& \multicolumn{1}{c|}{}                              & \multicolumn{1}{c|}{BN, Tanh} & -               & ($M \times n_b$)   & ($M \times n_b$)                 \\ \hline
				\multirow{6}{*}{$\mathcal{N}_{\gamma}$} & \multicolumn{1}{c|}{\multirow{3}{*}{MLP $2_1$}} & \multicolumn{1}{c|}{Linear}   & W($n_b, n_h$)             & ($K, n_b$)  & ($K, n_h$)                 \\ \cline{3-6} 
				& \multicolumn{1}{c|}{}                              & \multicolumn{1}{c|}{BN, Tanh} &  -                      & ($K, n_h$)   & ($K, n_h$)                 \\ \cline{3-6} 
				& \multicolumn{1}{c|}{}                              & \multicolumn{1}{c|}{Linear}   & W($n_h, n_z$)              & ($K, n_h$)   & ($K, n_z$)                 \\ \cline{2-6} 
				& \multicolumn{1}{c|}{\multirow{3}{*}{MLP $2_2$}} & \multicolumn{1}{c|}{Linear}   & W($n_b, n_h$)             & ($K, n_b$)  & ($K, n_h$)                 \\ \cline{3-6} 
				& \multicolumn{1}{c|}{}                              & \multicolumn{1}{c|}{BN, Tanh} &  -                      & ($K, n_h$)   & ($K, n_h$)                 \\ \cline{3-6} 
				& \multicolumn{1}{c|}{}                              & \multicolumn{1}{c|}{Linear}   & W($n_h, n_z$)              & ($K, n_h$)   & ($K, n_z$)                 \\ \hline
				CVAE Decoder                 & \multicolumn{3}{c|}{See Table \ref{tab:component_cvae} }                                  & ($K \times J, C \times n_{dct}$), ($K \times n_z$) & ($K \times J, C \times n_{dct}$) \\ \bottomrule
			\end{tabular}
		}
	\end{table}

	\begin{table}[!t]
		\caption{The network structure of the employed CVAE. For Human3.6M, $J=17$. For HumanEva-I, $J=15$. For both datasets, $C=3$, $F=256$, $n_{dct}=10$, $n_z=64$.}
		\label{tab:component_cvae}
		\resizebox{0.45\textwidth}{!}{
			\begin{tabular}{c|c|c|c|c|c}
				\toprule
				Component   & Block                   & Layer    & Weight Size              & Input Size    & Output Size \\ \midrule
				\multirow{20}{*}{Encoder} & \multirow{18}{*}{GCN 1} & GCL      & A($J, J$), W($2 \times n_{dct},  F$)  & ($J, 2 \times C \times n_{dct}$)       & ($J, F$)    \\ \cline{3-6}
				&                         & BN, Tanh &  -                     & ($J, F$)      & ($J, F$)    \\ \cline{3-6}
				&                         & GCL      & A($J, J$), W($F, F$)   & ($J, F$)      & ($J, F$)    \\ \cline{3-6}
				&                         & BN, Tanh &   -                     & ($J, F$)      & ($J, F$)    \\ \cline{3-6}
				&                         & GCL      & A($J, J$), W($F, F$)   & ($J, F$)      & ($J, F$)    \\ \cline{3-6}
				&                         & BN, Tanh &   -                     & ($J, F$)      & ($J, F$)    \\ \cline{3-6}
				&                         & GCL      & A($J, J$), W($F, F$)   & ($J, F$)      & ($J, F$)    \\ \cline{3-6}
				&                         & BN, Tanh &   -                     & ($J, F$)      & ($J, F$)    \\ \cline{3-6}
				&                         & GCL      & A($J, J$), W($F, F$)   & ($J, F$)      & ($J, F$)    \\ \cline{3-6}
				&                         & BN, Tanh &   -                     & ($J, F$)      & ($J, F$)    \\ \cline{3-6}
				&                         & GCL      & A($J, J$), W($F, F$)   & ($J, F$)      & ($J, F$)    \\ \cline{3-6}
				&                         & BN, Tanh &    -                    & ($J, F$)      & ($J, F$)    \\ \cline{3-6}
				&                         & GCL      & A($J, J$), W($F, F$)   & ($J, F$)      & ($J, F$)    \\ \cline{3-6}
				&                         & BN, Tanh &  -                    & ($J, F$)      & ($J, F$)    \\ \cline{3-6}
				&                         & GCL      & A($J, J$), W($F, F$)   & ($J, F$)      & ($J, F$)    \\ \cline{3-6}
				&                         & BN, Tanh &  -                     & ($J, F$)      & ($J, F$)    \\ \cline{3-6}
				&                         & GCL      & A($J, J$), W($F, F$)   & ($J, F$)      & ($J, F$)    \\ \cline{3-6}
				&                         & BN, Tanh & -                    & ($J, F$)      & ($J, F$)    \\ \cline{2-6} 
				& MLP 1                   & Linear & W($J \times F, n_z$)             & ($J \times F$)       & $(n_z)$         \\ \cline{3-6} \cline{2-6} 
				& MLP 2                   & Linear & W($J \times F, n_z$)             & ($J \times F$)       & $(n_z)$         \\ \hline
				\multirow{19}{*}{Decoder} & \multirow{19}{*}{GCN 2} & GCL      & A($J, J$), W($C \times n_{dct} + n_z,  F$) & ($J, C \times n_{dct}$), ($n_z$) & ($J, F$)    \\ \cline{3-6}
				&                         & BN, Tanh &  -                     & ($J, F$)      & ($J, F$)    \\ \cline{3-6}
				&                         & GCL      & A($J, J$), W($F, F$)   & ($J, F$)      & ($J, F$)    \\ \cline{3-6}
				&                         & BN, Tanh &  -                    & ($J, F$)      & ($J, F$)    \\ \cline{3-6}
				&                         & GCL      & A($J, J$), W($F, F$)   & ($J, F$)      & ($J, F$)    \\ \cline{3-6}
				&                         & BN, Tanh &  -                     & ($J, F$)      & ($J, F$)    \\ \cline{3-6}
				&                         & GCL      & A($J, J$), W($F, F$)   & ($J, F$)      & ($J, F$)    \\ \cline{3-6}
				&                         & BN, Tanh &  -                    & ($J, F$)      & ($J, F$)    \\ \cline{3-6}
				&                         & GCL      & A($J, J$), W($F, F$)   & ($J, F$)      & ($J, F$)    \\ \cline{3-6}
				&                         & BN, Tanh &  -                     & ($J, F$)      & ($J, F$)    \\ \cline{3-6}
				&                         & GCL      & A($J, J$), W($F, F$)   & ($J, F$)      & ($J, F$)    \\ \cline{3-6}
				&                         & BN, Tanh & -                       & ($J, F$)      & ($J, F$)    \\ \cline{3-6}
				&                         & GCL      & A($J, J$), W($F, F$)   & ($J, F$)      & ($J, F$)    \\ \cline{3-6}
				&                         & BN, Tanh &   -                     & ($J, F$)      & ($J, F$)    \\ \cline{3-6}
				&                         & GCL      & A($J, J$), W($F, F$)   & ($J, F$)      & ($J, F$)    \\ \cline{3-6}
				&                         & BN, Tanh &   -                     & ($J, F$)      & ($J, F$)    \\ \cline{3-6}
				&                         & GCL      & A($J, J$), W($F, F$)   & ($J, F$)      & ($J, F$)    \\ \cline{3-6}
				&                         & BN, Tanh &  -                   & ($J, F$)      & ($J, F$)    \\ \cline{3-6}
				&                         & GCL      & A($J, J$), W($F, C \times n_{dct}$)    & ($J, F$)      & ($J, C \times n_{dct}$)     \\ 
				\bottomrule
		\end{tabular}}
	\end{table}

	Now, we introduce the Graph Convolutional Network (GCN). As shown in Table \ref{tab:component_ours}, our GCN is made up of five sequentially stacked GCL-BN-Tanh layers, where GCL stands for Graph Convolutional layer. Let $\textbf{H}^l\in{\mathbb{R}^{J\times F^l}}$ be the input to the $l^{th}$ GCL where $F^{l}$ is the hidden feature dimension size, $\textbf{A}^l\in{\mathbb{R}^{J\times J}}$ the adjacency matrix, and $\textbf{W}^l\in{\mathbb{R}^{F^l\times F^{l+1}}}$ the trainable parameters, the GCL executes the following computation:
	\begin{equation}
		\textbf{H}^{l+1} = \textbf{A}^l\textbf{H}^l\textbf{W}^l,
	\end{equation}
	where $\textbf{H}^{l+1}\in{\mathbb{R}^{J\times F^{l+1}}}$ is the output of the $l^{th}$ GCL. At the very beginning, $F^0=C\times n_{dct}$.

	\begin{figure*}[!t]
		\centering
		\includegraphics[width=\textwidth]{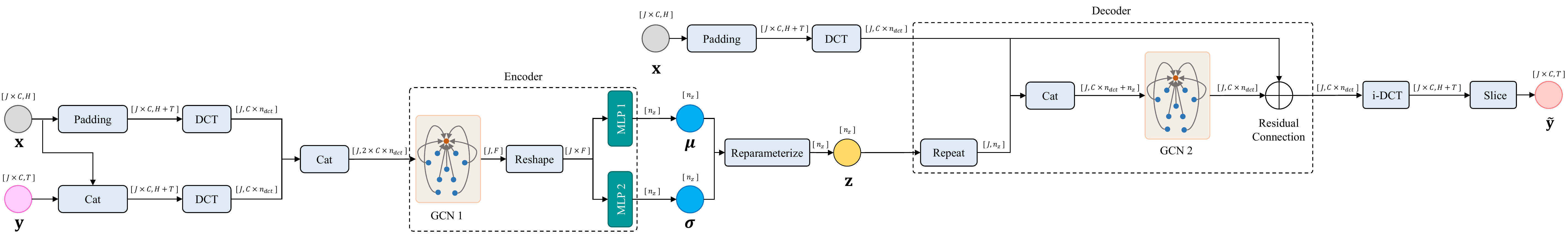}\\
		\caption{Detailed architecture of the adopted CVAE model. It is composed of an encoder and a decoder that are built on GCNs and MLPs. Please refer to the main text for detailed descriptions of the architecture.}
		\label{fig:cvae-arch}
	\end{figure*}

	\section{CVAE Network Architecture}
	Recall that before applying our method to generate diverse results, we need to train a CVAE model beforehand. The CVAE network architecture adopted in this paper is shown in Figure \ref{fig:cvae-arch}. 
	Let $\mathbf{x} \in \mathbb{R}^{[J \times C, H]}$ be an observed pose sequence, and $\mathbf{y} \in \mathbb{R}^{[J \times C, T]}$ be the ground truth future poses. We compute the frequency coefficients of shape $[J, C \times n_{dct}]$ from each of them and then concatenate both the frequency content into data of shape $[J, 2 \times C \times n_{dct}]$.
	Then, an encoder is used to learn the parameters of the posterior Gaussian distribution $\mathcal{N}(\bm{\mu}, \bm{\sigma})$ of the latent code $\mathbf{z}$ given $\mathbf{x}$ and $\mathbf{y}$, where $\bm{\mu} \in \mathbb{R}^{[n_z]}$ is the mean of the posterior distribution, and $\bm{\sigma} \in \mathbb{R}^{[n_z]}$ is the diagonal values of the co-variance matrix of the posterior distribution.
	Particularly, as shown in Table \ref{tab:component_cvae}, the encoder consists of a GCN and two MLPs. The GCN is composed of nine GCL-BN-Tanh layers to extract hidden features of shape $[J, F]$. The two MLPs, each of which just comprises a single Linear layer, map the hidden features into $\bm{\mu}$ and $\bm{\sigma}$, respectively.
	Then a latent variable $\mathbf{z} \in \mathbb{R}^{[n_z]}$ can be drawn from the posterior Gaussian distribution by the reparameterization trick.
	
	Next, a decoder is used to reconstruct $\mathbf{y}$ from the latent code $\mathbf{z}$.
	To achieve that, we first repeat $\mathbf{z}$, $J$ times and concatenate them with the DCT coefficients of $\mathbf{x}$, resulting in a feature of shape $[J, C \times n_{dct} + n_z]$.
	Afterwards, we employ another GCN comprising 9 GCL-BN-Tanh layers to extract hidden features of shape $[J, F]$, and a GCL layer that projects the hidden features back into the frequency space of shape $[J, C \times n_{dct}]$.
	The frequency coefficients of $\mathbf{x}$ after DCT is added to this output.
	Finally, we use an i-DCT function to transform the above output back into the pose space and slice off the future $T$ frames to obtain $\mathbf{\tilde{y}} \in \mathbb{R}^{[J \times C, T]}$.
	
	
	\begin{table*}[!t]
		\caption{Comparison on four additional metrics: ADE-m, FDE-m, FID, and ACC.}
		\label{tab:other_matrics}
		\resizebox{0.7\textwidth}{!}{
			\begin{tabular}{c|cccc|cccc}
				\hline
				& \multicolumn{4}{c|}{Human3.6M \cite{ionescu2013human3}} & \multicolumn{4}{c}{HumanEva-I \cite{sigal2010humaneva}}                                                                   \\ \cline{2-9} 
				&  ADE-m $\downarrow$ & FDE-m $\downarrow$ & FID $\downarrow$ & ACC $\uparrow$ &  ADE-m $\downarrow$ & FDE-m $\downarrow$ & FID $\downarrow$ & ACC $\uparrow$\\ \hline
				DLow \cite{yuan2020dlow} & \textbf{0.896}   & \textbf{1.284}  & \textbf{1.566}    & 0.227  & \textbf{0.577}   & \textbf{0.717}  & 3.472    & 0.527 \\
				GSPS \cite{mao2021generating} & 1.013   & 1.372  & 1.915    & 0.222  & 0.686   & 0.794  & 1.604    & 0.516 \\
				Ours & 0.924   & 1.344  & 2.060    & \textbf{0.261}    & 0.716   & 0.770  & \textbf{1.106}    &  \textbf{0.609}\\ \hline
			\end{tabular}
		}
	\end{table*}
	
	\begin{table*}[!t]
		\caption{Comparison with DLow \cite{yuan2020dlow} when adding random noises of different variance to its Gaussian distributions.}
		\label{tab:noise_onto_dlow}
		\resizebox{1\textwidth}{!}{
			\begin{tabular}{c|cccccccc|cccccccc}
				\toprule
				\multirow{3}{*}{} & \multicolumn{8}{c|}{Human3.6M \cite{ionescu2013human3}}                                                                                                                                                   & \multicolumn{8}{c}{HumanEva-I \cite{sigal2010humaneva}}                                                                                                                                                  \\ \cline{2-17} 
				& \multicolumn{1}{c|}{\multirow{2}{*}{Ours}} & \multicolumn{1}{c|}{\multirow{2}{*}{DLow \cite{yuan2020dlow}}} & \multicolumn{3}{c|}{DLow-variant \cite{yuan2020dlow} w/o retraining}           & \multicolumn{3}{c|}{DLow-variant \cite{yuan2020dlow} w/ retraining} & \multicolumn{1}{c|}{\multirow{2}{*}{Ours}} & \multicolumn{1}{c|}{\multirow{2}{*}{DLow \cite{yuan2020dlow}}} & \multicolumn{3}{c|}{DLow-variant \cite{yuan2020dlow} w/o retraining}           & \multicolumn{3}{c}{DLow-variant \cite{yuan2020dlow} w/ retraining} \\ \cline{4-9} \cline{12-17} 
				& \multicolumn{1}{c|}{}                      & \multicolumn{1}{c|}{}                      & $\sigma$ = 1 & $\sigma$ = 1.7 & \multicolumn{1}{c|}{$\sigma$ = 2} & $\sigma$ = 1     & $\sigma$ = 1.7    & $\sigma$ = 2    & \multicolumn{1}{c|}{}                      & \multicolumn{1}{c|}{}                      & $\sigma$ = 1 & $\sigma$ = 5.7 & \multicolumn{1}{c|}{$\sigma$ = 6} & $\sigma$ = 1    & $\sigma$ = 5.7    & $\sigma$ = 6    \\ \hline
				APD $\uparrow$              & \multicolumn{1}{c|}{15.310}                & \multicolumn{1}{c|}{11.741}                & 15.190 & 19.295   & \multicolumn{1}{c|}{\textbf{20.894}} & 11.100     & 15.373      & 18.667    & \multicolumn{1}{c|}{6.109}                 & \multicolumn{1}{c|}{4.855}                 & 4.753  & 6.147    & \multicolumn{1}{c|}{\textbf{6.243}}  & 4.488     & 6.135       & 6.205     \\
				ADE $\downarrow$              & \multicolumn{1}{c|}{\textbf{0.370}}                 & \multicolumn{1}{c|}{0.425}                 & 0.560  & 0.721    & \multicolumn{1}{c|}{0.795}  & 0.526      & 0.675       & 0.748     & \multicolumn{1}{c|}{\textbf{0.220}}                 & \multicolumn{1}{c|}{0.251}                 & 0.305  & 0.647    & \multicolumn{1}{c|}{0.654}  & 0.297     & 0.638       & 0.649     \\
				FDE $\downarrow$              & \multicolumn{1}{c|}{\textbf{0.485}}                 & \multicolumn{1}{c|}{0.518}                 & 0.674  & 0.863    & \multicolumn{1}{c|}{0.949}  & 0.627      & 0.799       & 0.883     & \multicolumn{1}{c|}{\textbf{0.234}}                 & \multicolumn{1}{c|}{0.268}                 & 0.327  & 0.658    & \multicolumn{1}{c|}{0.664}  & 0.321     & 0.648       & 0.657     \\
				MMADE $\downarrow$             & \multicolumn{1}{c|}{\textbf{0.475}}                 & \multicolumn{1}{c|}{0.495}                 & 0.612  & 0.764    & \multicolumn{1}{c|}{0.835}  & 0.579      & 0.719       & 0.789     & \multicolumn{1}{c|}{\textbf{0.342}}                 & \multicolumn{1}{c|}{0.362}                 & 0.387  & 0.659    & \multicolumn{1}{c|}{0.666}  & 0.381     & 0.652       & 0.664     \\
				MMFDE $\downarrow$             & \multicolumn{1}{c|}{\textbf{0.516}}                 & \multicolumn{1}{c|}{0.531}                 & 0.682  & 0.868    & \multicolumn{1}{c|}{0.953}  & 0.634      & 0.804       & 0.889     & \multicolumn{1}{c|}{\textbf{0.316}}                 & \multicolumn{1}{c|}{0.339}                 & 0.372  & 0.661    & \multicolumn{1}{c|}{0.667}  & 0.368     & 0.650       & 0.661     \\ \bottomrule
		\end{tabular}}
	\end{table*}

	\begin{table*}[!t]
		\caption{Comparison with learning $\pi$ (in Eq. 12) automatically.}
		\label{tab:learned_pi}
		\resizebox{1\textwidth}{!}{
			\begin{tabular}{c|ccccc|ccccc}
				\toprule
				\multirow{2}{*}{}      & \multicolumn{5}{c|}{Human3.6M \cite{ionescu2013human3}}         & \multicolumn{5}{c}{HumanEva-I \cite{sigal2010humaneva}}        \\ \cline{2-11} 
				& APD $\uparrow$   & ADE$\downarrow$   & FDE $\downarrow$  & MMADE $\downarrow$ & MMFDE $\downarrow$ & APD $\uparrow$  & ADE $\downarrow$  & FDE $\downarrow$  & MMADE $\downarrow$ & MMFDE $\downarrow$ \\ \hline
				DLow \cite{yuan2020dlow}     & 11.741 & 0.425 & 0.518 & 0.495 & 0.531 & 4.855 & 0.251 & 0.268 & 0.362 & 0.339 \\
				GSPS  \cite{mao2021generating}    & 14.757 & 0.389 & 0.496 & 0.476 & 0.525 & 5.825 & 0.233 & 0.244 & 0.343 & 0.331 \\
				Ours                         & \textbf{15.310} & 0.370 & \textbf{0.485} & \textbf{0.475} & \textbf{0.516} & \textbf{6.109} & \textbf{0.220} & \textbf{0.234} & 0.342 & 0.316 \\
				Ours-Individual-$\pi$ & 13.530 & 0.372 & 0.493 & 0.481 & 0.525 & 5.606 & 0.229 & 0.239 & 0.338 & 0.317 \\
				Ours-Shared-$\pi$     & 14.440 & \textbf{0.368} & \textbf{0.485} & 0.477 & 0.519 & 5.690 & 0.228 & 0.235 & \textbf{0.324} & \textbf{0.295} \\ \bottomrule
		\end{tabular}}
	\end{table*}

	\section{Evaluation on FID, ACC, ADE-m, and FDE-m}
	
	We further evaluate our method on four additional metrics: FID, ACC, ADE-median (ADE-m), and FDE-median (FDE-m).
	
	\begin{itemize}
		\item Recognition Accuracy (\textbf{ACC}). A pre-trained action recognition classifier is used to classify the generated poses. ACC is the overall recognition accuracy. We train the action classifier in the way suggested by~\cite{bie2022hit}.
		\item Frechet Inception Distance (\textbf{FID}). Features are extracted from the generated and real data by the pre-trained action classifier. FID is then calculated as the Frechet inception distance between the two feature distributions.
		\item \textbf{ADE-m} and \textbf{FDE-m} are similar to ADE and FDE except that the median distance are reported.
	\end{itemize}

	
	
	The results on the four metrics are shown in Table~\ref{tab:other_matrics}.
	For ADE-m and FDE-m, DLow performs the best. That is because the diversity by DLow (11.741, Human3.6M) is much lower than that by GSPS (14.757) and our method (15.310). The lower the diversity, the lower the median distance. Therefore, it is not surprise that both our method and GSPS have larger ADE-m and FDE-m than DLow. Compared with GSPS, our method has lower ADE-m and FDE-m (ADE-m: 0.924 (our) \textit{v.s.} 1.013 (GSPS), FDE-m: 1.344 (our) \textit{v.s.} 1.372 (GSPS), Human3.6M), even though our method has greater diversity (APD: 15.310 (our) \textit{v.s.} 14.757 (GSPS)).
	
	For FID and ACC, our method is better than DLow and GSPS in terms of ACC. For HumanEva-I, our FID is the best. However, an exception is the FID of Human3.6M, for which DLow performs much better than GSPS and our method. Again, this is because the diversity of the results of DLow (APD=11.741) is much lower than those of GSPS (14.757) and our method (15.310).
	
	
	\section{Comparison with a variant of DLow}
	
	We simply add some random noise to the generated Gaussian distributions (i.e., adding noises to the mean and variance of the Gaussian distributions) in DLow \cite{yuan2020dlow} and compare with this variant of DLow.
	
	
	
	Firstly, we add noises to the Gaussian distributions generated by DLow without re-training the DLow model (DLow-variant w/o retraining). Secondly, we retrain the DLow model, and add noises to the Gaussian distributions at both training and testing phases (DLow-variant w/ retraining). The noises are randomly drawn from a Normal distribution $\mathcal{N}(0,\sigma)$. The mean of the noises is always zero, while for comparison we test noises of different variances.
	
	As shown in Table~\ref{tab:noise_onto_dlow}, adding noises can increase the diversity of the generated results (measured by APD). The heavier the noises (produced by larger $\sigma$), the more diverse the results. However, the negative effect is that the accuracy of the results is decreased. For the Human3.6M dataset, please see the columns of ``DLow-variant w/o retraining $\sigma=1$'' and ``DLow-variant w/ retraining $\sigma=1.7$'' that produce similar APD as ours. Their accuracy metrics are much higher than ours. Although adding noises can yield very large APD (20.894 in the column of ``DLow-variant w/o retraining $\sigma=2$''), the generated poses look unrealistically. And those results on the HumanEva-I dataset have the same trend. From this point of view, our method is better than directly adding noises to generated Gaussian distributions.
	
	\section{Learn $\pi$ instead of setting a constant value}
	In the main paper, we set $\pi$ (see Eq. 12) to a constant value. With constant $\pi$ (1/40=0.025), each basis vector has the equal probability to be assigned with the highest weight among all the basis vectors. To treat all the basis vectors equally, we therefore use the same constant $\pi$ (actually a probability) for each of them.
	
	As a variant, $\pi$ can also be learned automatically. We conduct experiments to compare between learning and setting a constant $\pi$. 
	The results are shown in Table~\ref{tab:learned_pi}.

	In the first experiment, we learn a $\pi$ for each input sample individually (Ours-Individual-$\pi$), the results are slightly worse than those of directly indicating a constant $\pi$. In the second experiment, we learn a $\pi$ shared by all the input samples (Ours-Shared-$\pi$), the new results are comparable to those of constant $\pi$. We find that the values of the learned shared $\pi$ fall in the range of [0.021, 0.029], which are nearly equally distributed.

	\begin{figure*}[!t]
		\flushleft \hspace{5.7cm} Human3.6M \hspace{4.4cm} HumanEva-I \\
		\centering
		\includegraphics[width=0.65\linewidth]{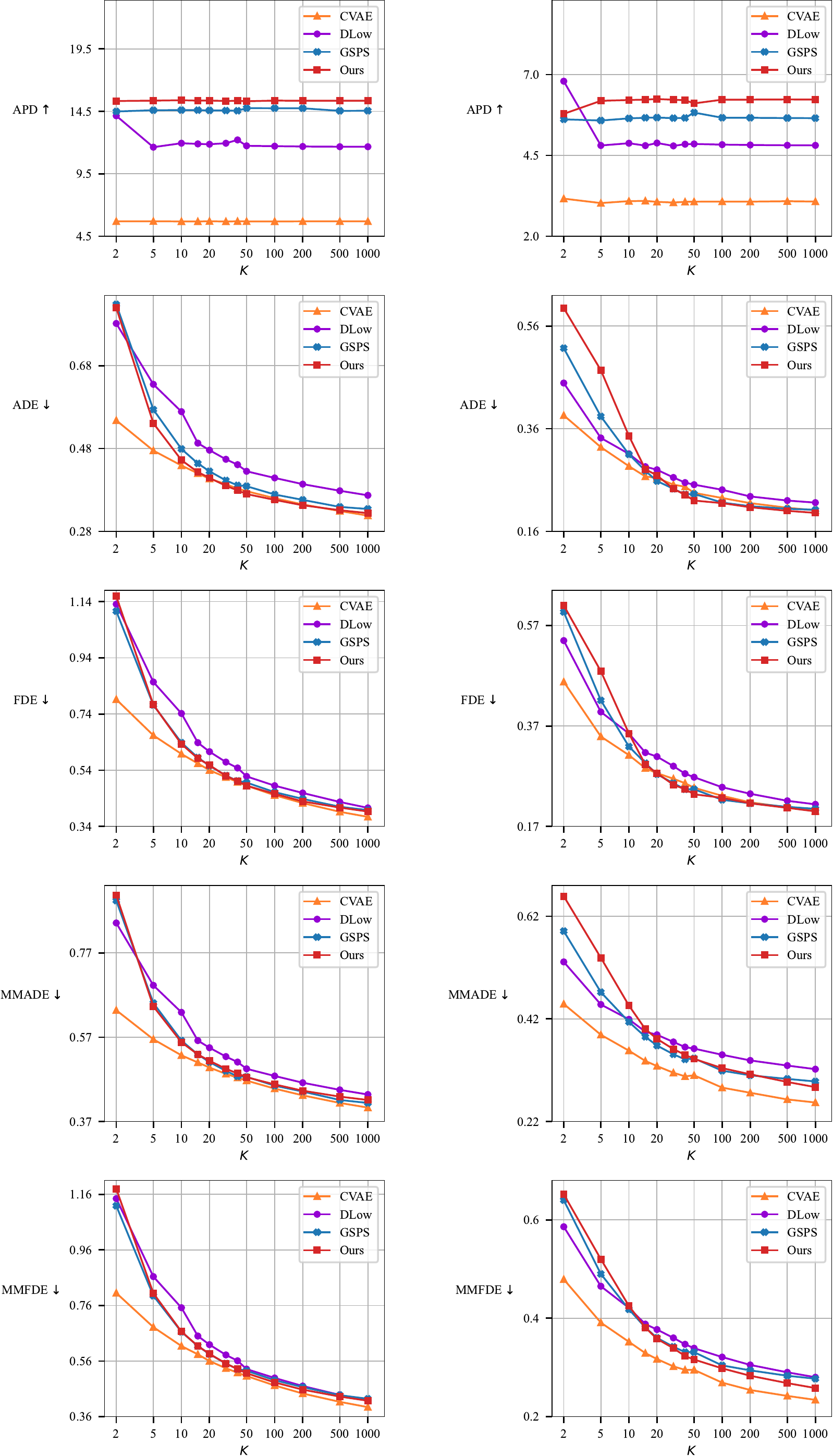} \\
		\caption{The first row shows how APD of different methods (including CVAE, DLow, GSPS, and our method) varies as $K$ increases. The other rows show the trends of ADE, FDE, MMADE, MMFDE, respectively. The figures on the left are plotted based on the data computed on Human3.6M, while those on the right are plotted based on HumanEva-I.}
		\label{fig:samples_K}
	\end{figure*}
	
	\section{More ablation studies on parameter $K$}
	
	In the main paper, we have demonstrated that $K$, \textit{i.e.}, the number of predictions for an input, has a large effect on the accuracy but not the diversity of the results produced by our method. Here, we show more ablation studies on $K$, and perform comparisons between CVAE random sampling, DLow~\cite{yuan2020dlow}, GSPS~\cite{mao2021generating} and our method.
	
	The results are plotted in Figure~\ref{fig:samples_K}. The first row shows the predicted results' diversity measured by \textit{APD}. As can be seen, for all the compared methods, \textit{APD} does not change much as $K$ increases. We can also see that the diversity of our results is the largest among all the methods, while that of CVAE is the smallest.
	
	The second and third rows show the predicted results' accuracy measured by \textit{ADE} and \textit{FDE}. As can be seen, for all the compared methods, \textit{ADE} and \textit{FDE} decrease as $K$ increases, indicating that more accurate results are obtained. 
	
	The fourth and fifth rows show the predicted results' accuracy measured by \textit{MMADE} and \textit{MMFDE}. For all the compared methods, \textit{MMADE} and \textit{MMFDE} decreases too as $K$ increases. 
	
	Observing all the results in Figure~\ref{fig:samples_K}, we can see that our method outputs more diverse results than DLow and GSPS, and at the same time our results are more accurate than those of DLow and GSPS. Generally, diversity and accuracy are two contradictory objectives. Low diversity usually means high accuracy. That is why CVAE, which generates results of the lowest diversity, produces the most accurate results in the second to fifth rows.


	\section{More Qualitative Comparisons}
	
	\begin{figure*}[!t]
		\centering
		\includegraphics[width=0.49\linewidth]{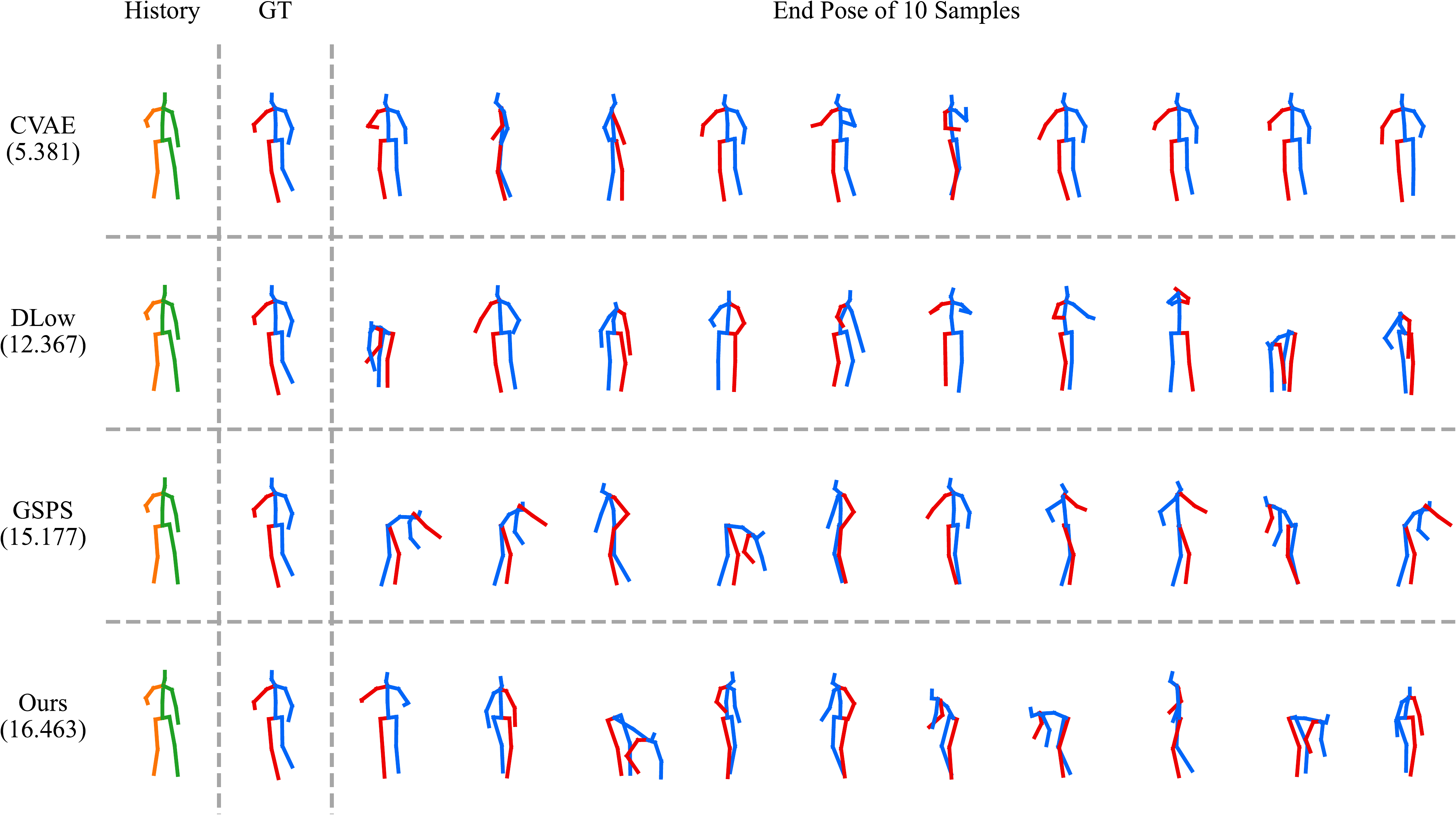}
		\includegraphics[width=0.49\linewidth]{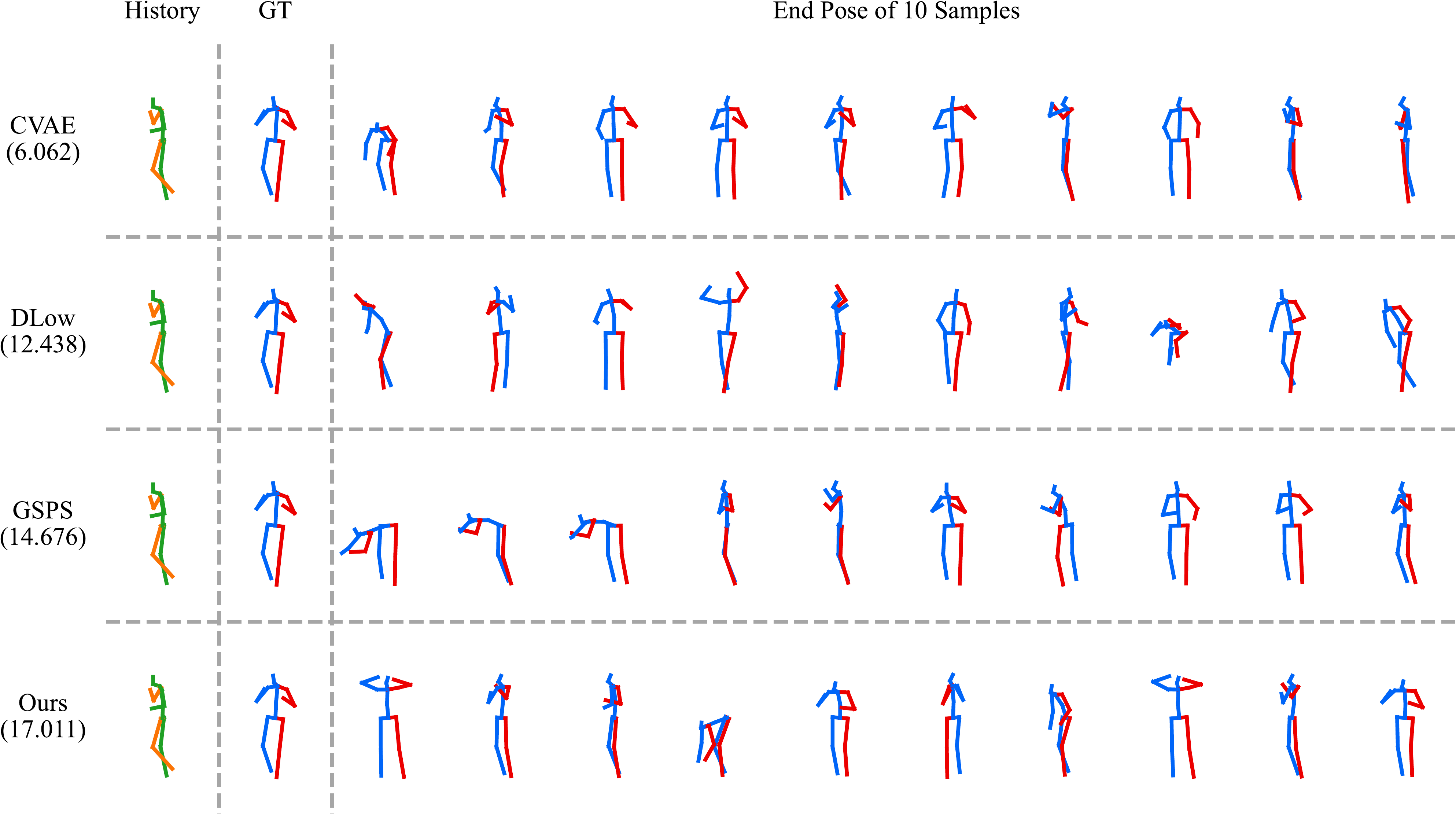}\\
		\includegraphics[width=0.49\linewidth]{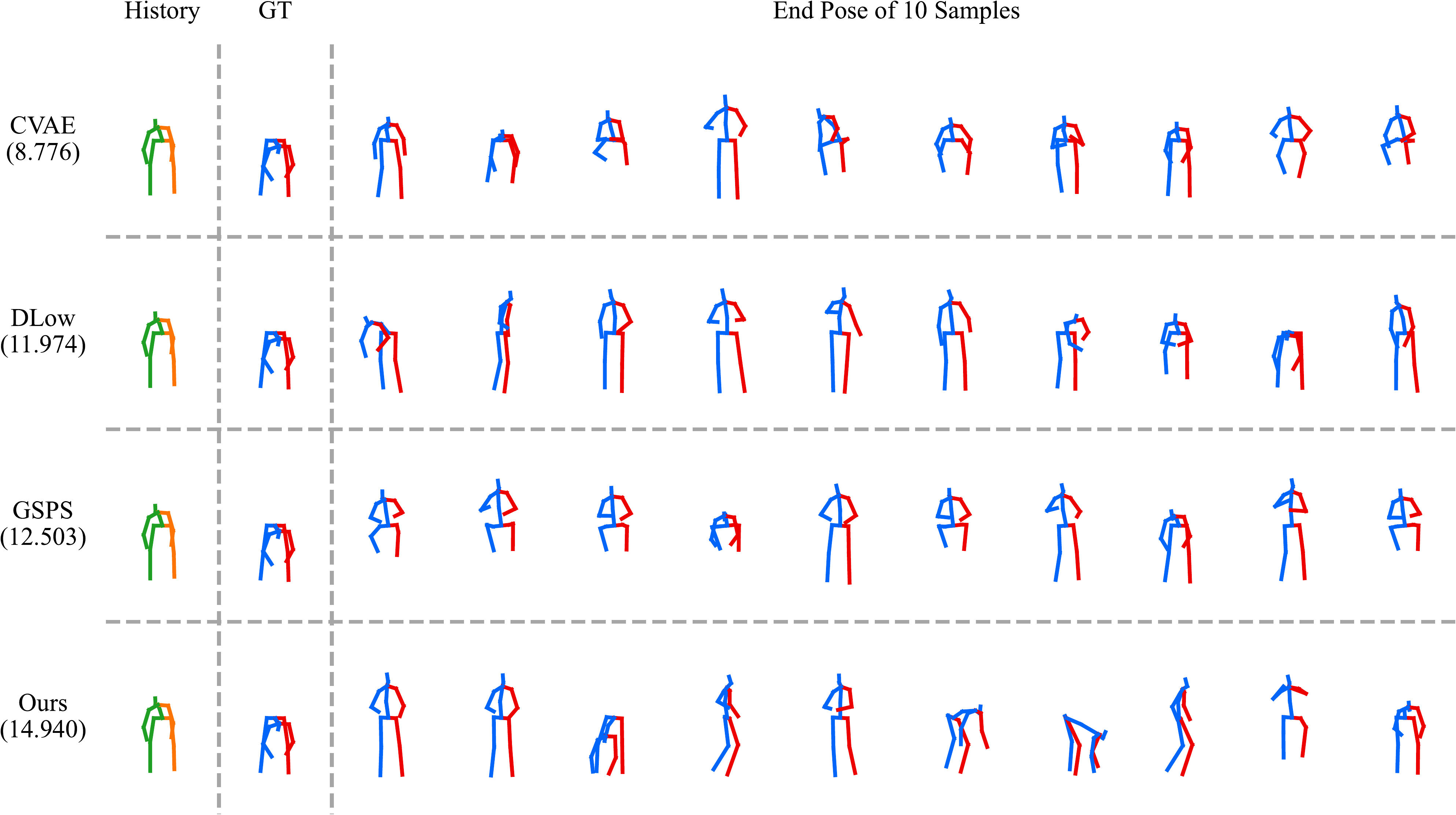}
		\includegraphics[width=0.49\linewidth]{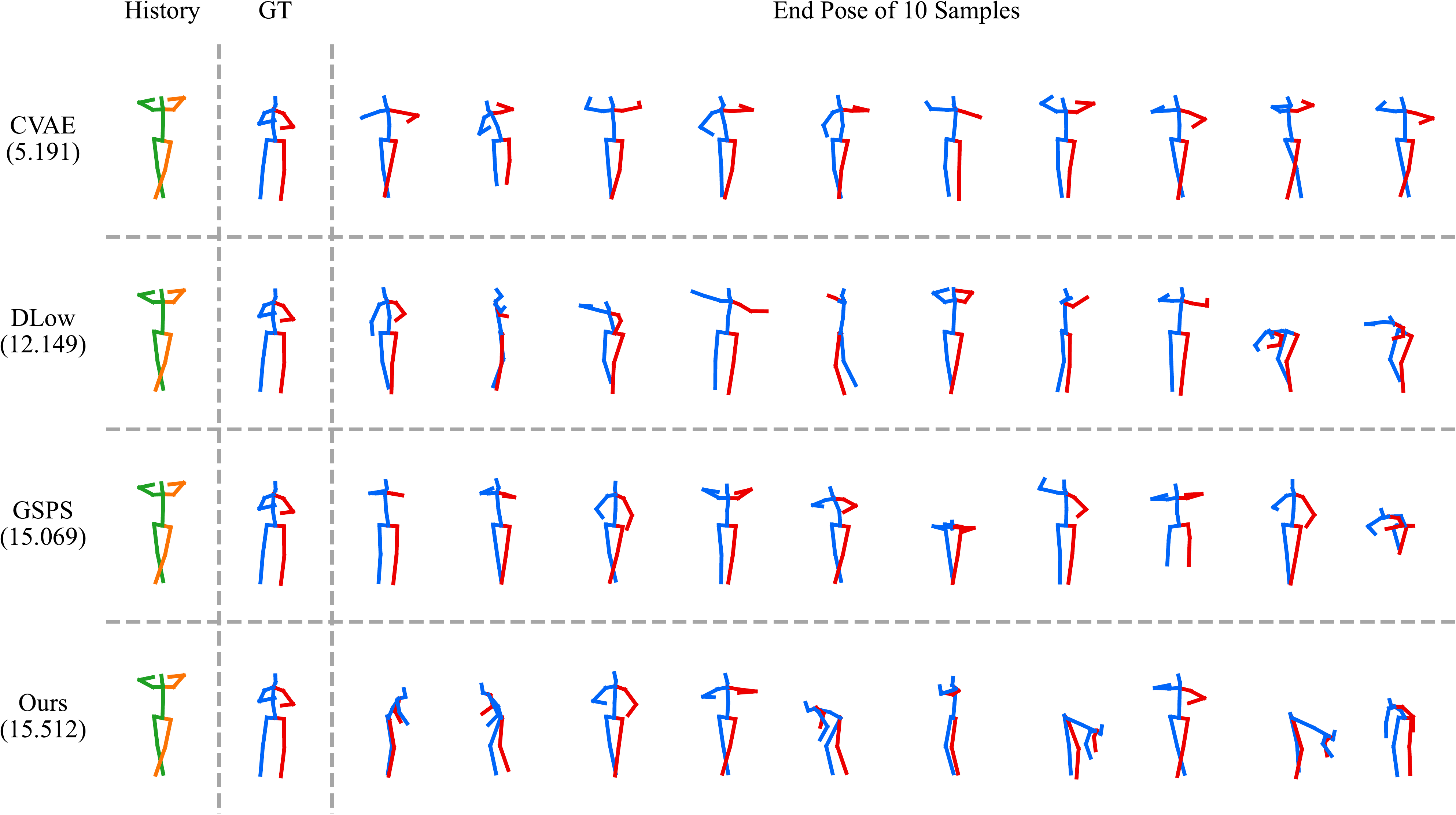} \\
		(a) Human3.6M \\
		\includegraphics[width=0.49\linewidth]{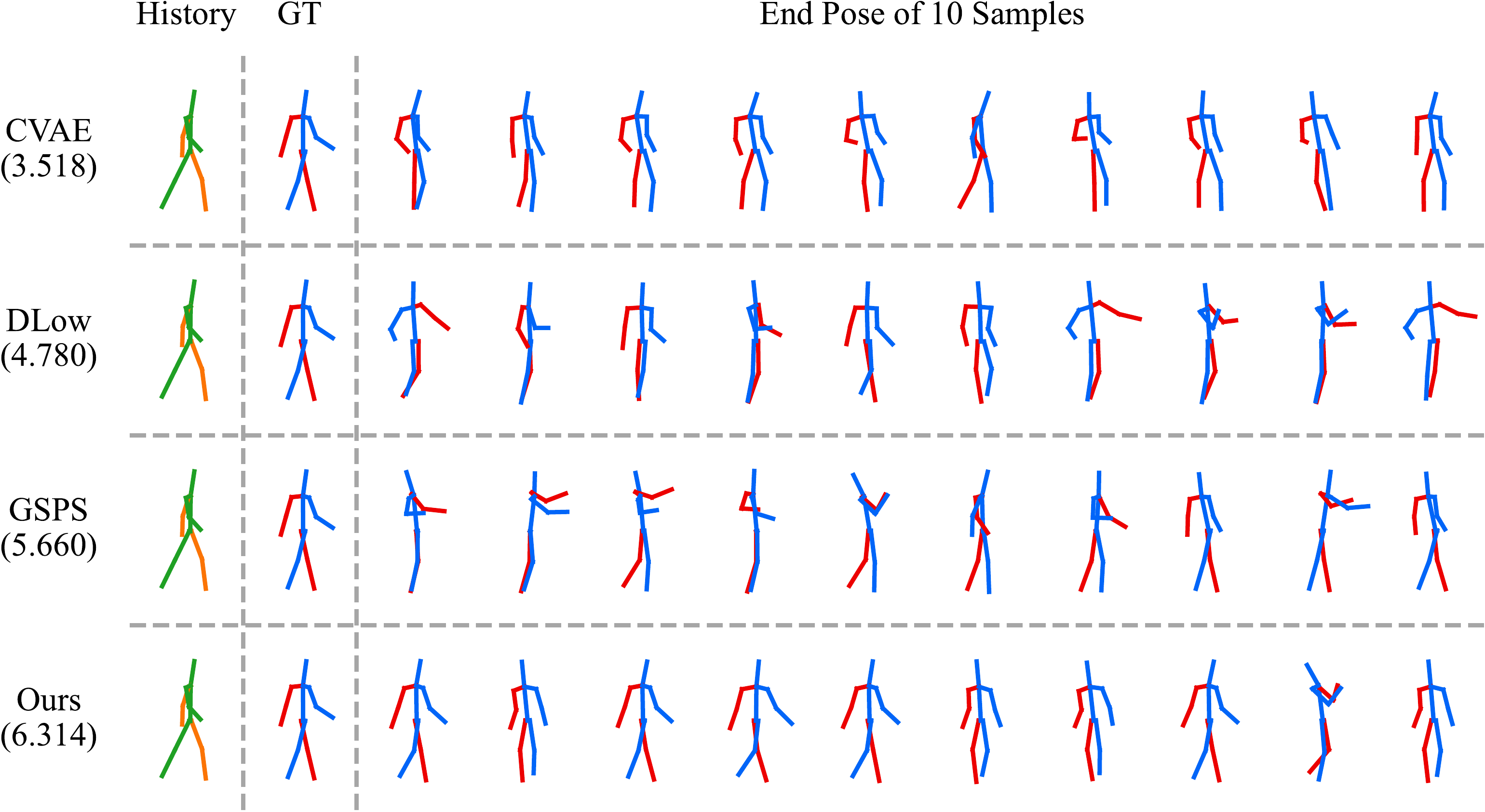}
		\includegraphics[width=0.49\linewidth]{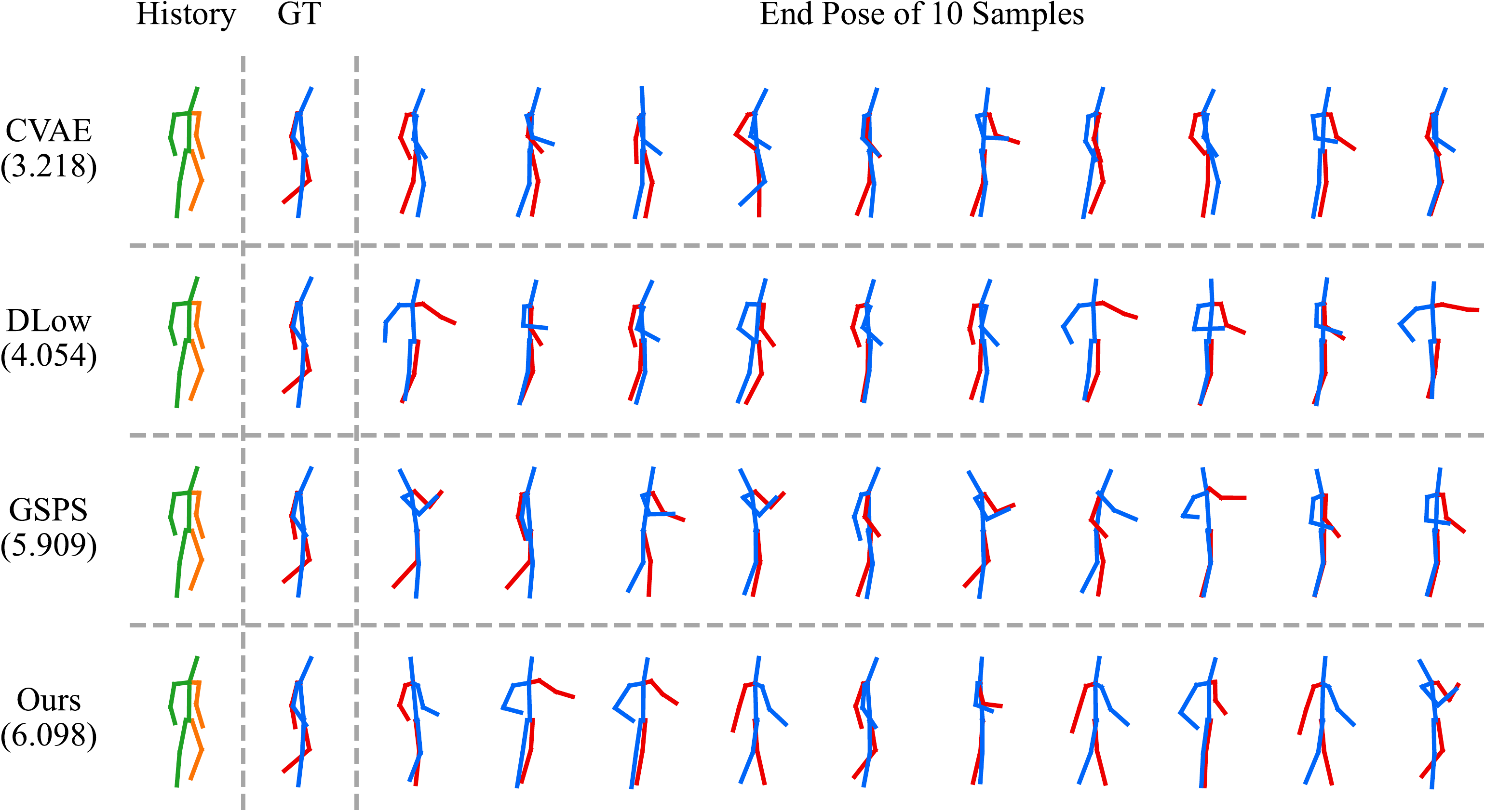} \\
		\includegraphics[width=0.49\linewidth]{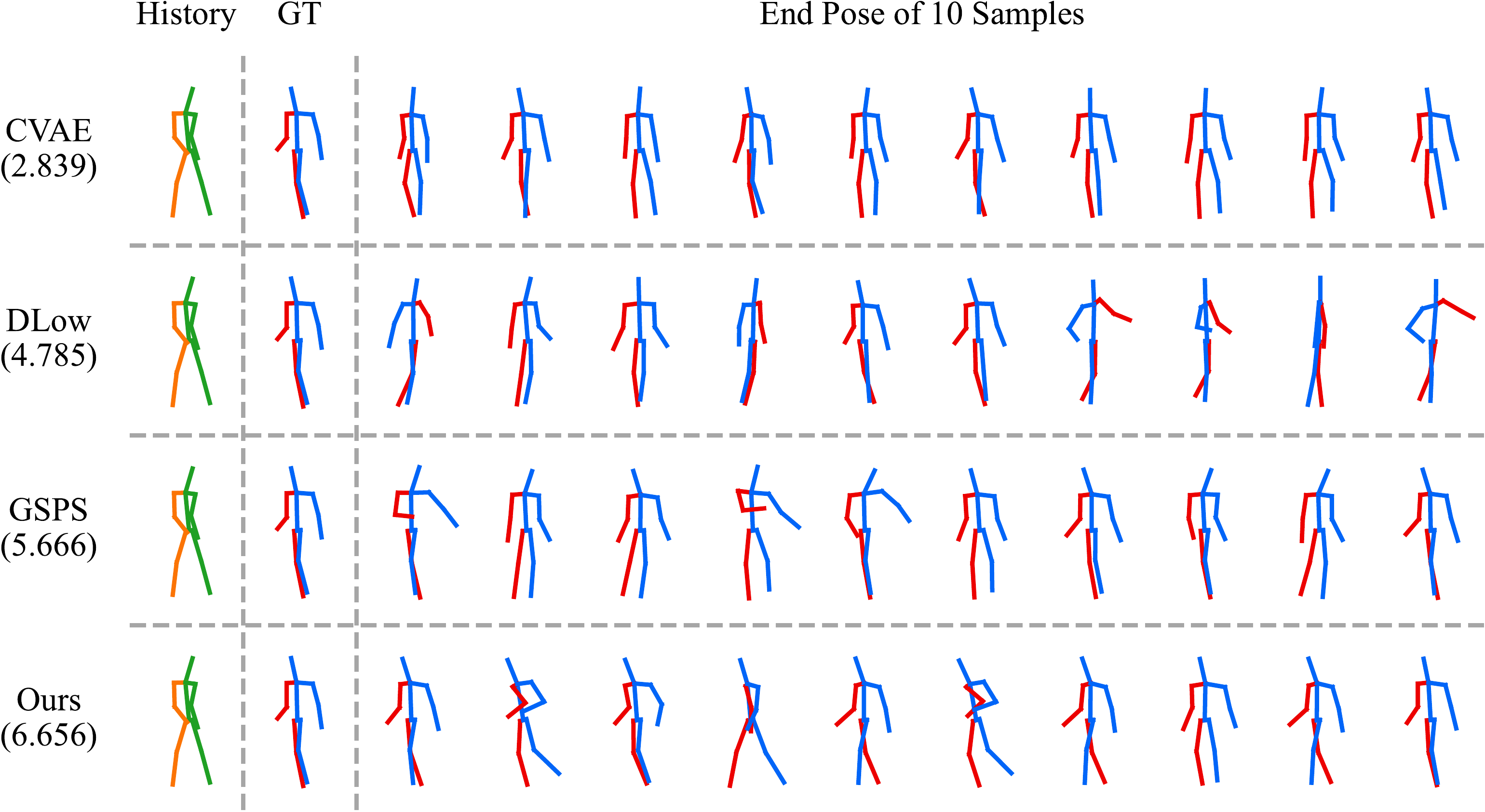}
		\includegraphics[width=0.49\linewidth]{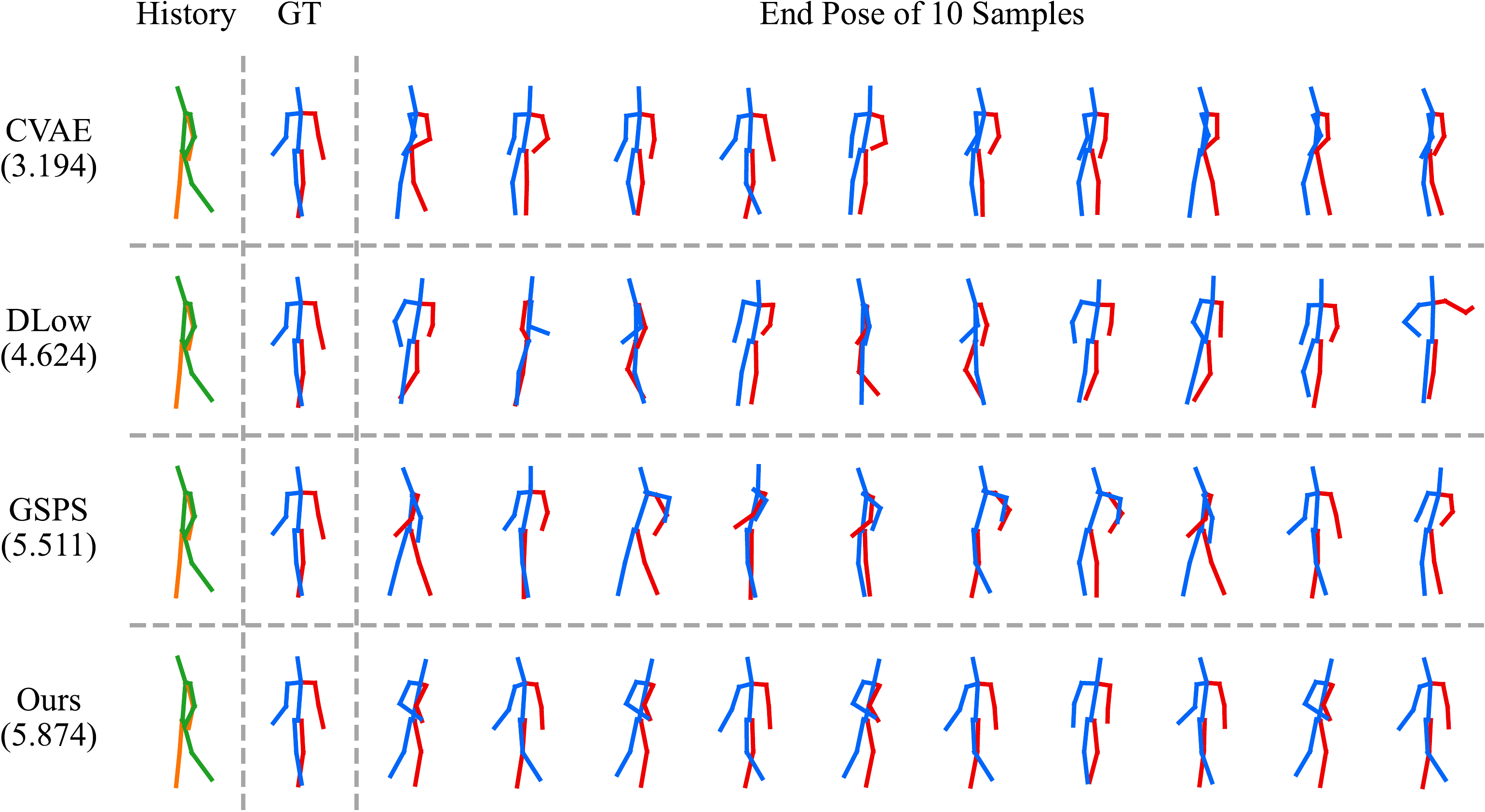}\\ 
		(b) HumanEva-I \\
		\caption{More qualitative results of CVAE, DLow, GSPS, and our method. The numbers in the brackets below the names of different methods show the diversity of the results computed by these methods. In these examples, our results are more diverse than the results of the other methods.}
		\label{fig:more-qual_1}
	\end{figure*}

	\begin{figure*}[!t]
		\centering
		\includegraphics[width=0.49\linewidth]{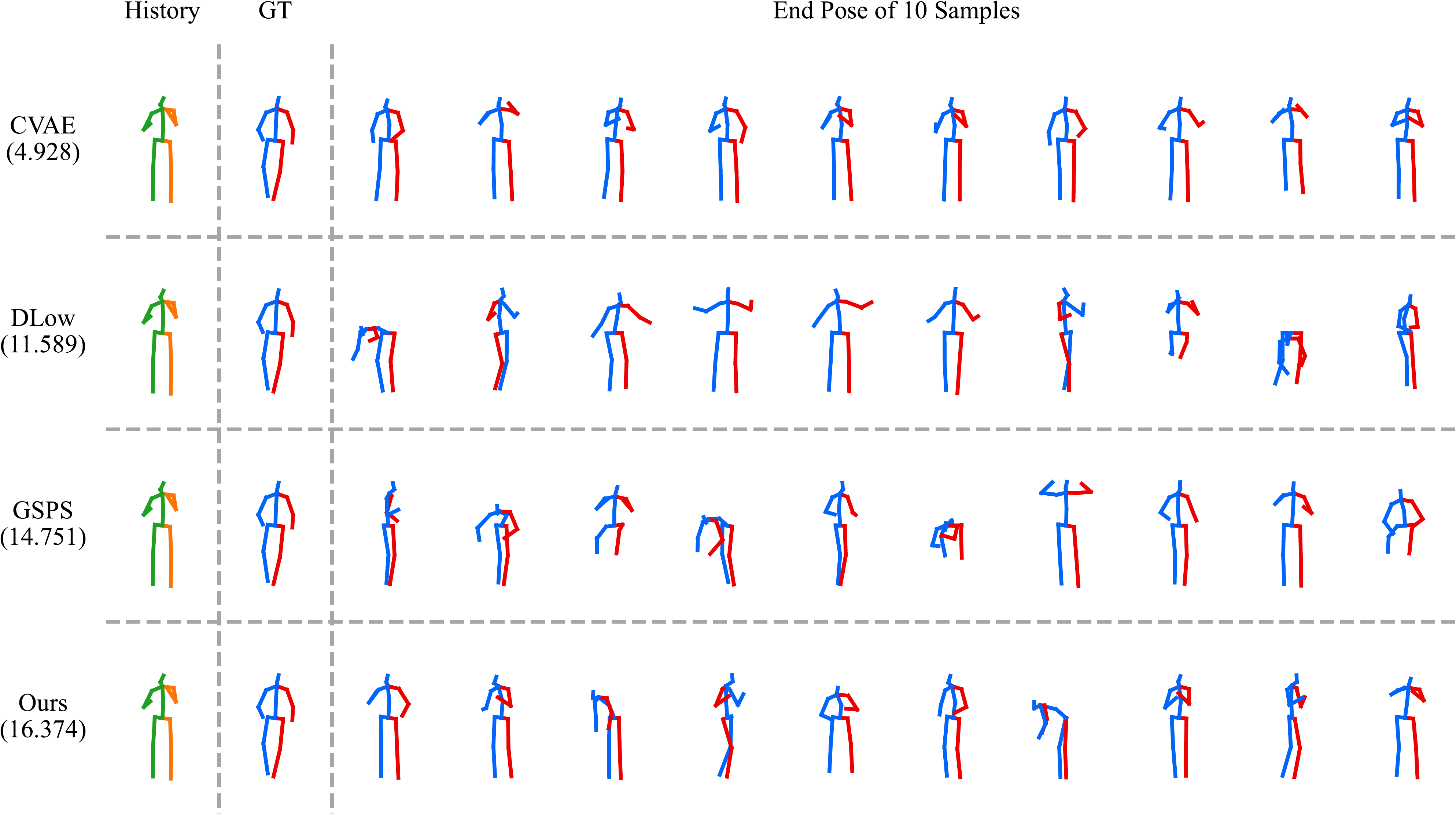}
		\includegraphics[width=0.49\linewidth]{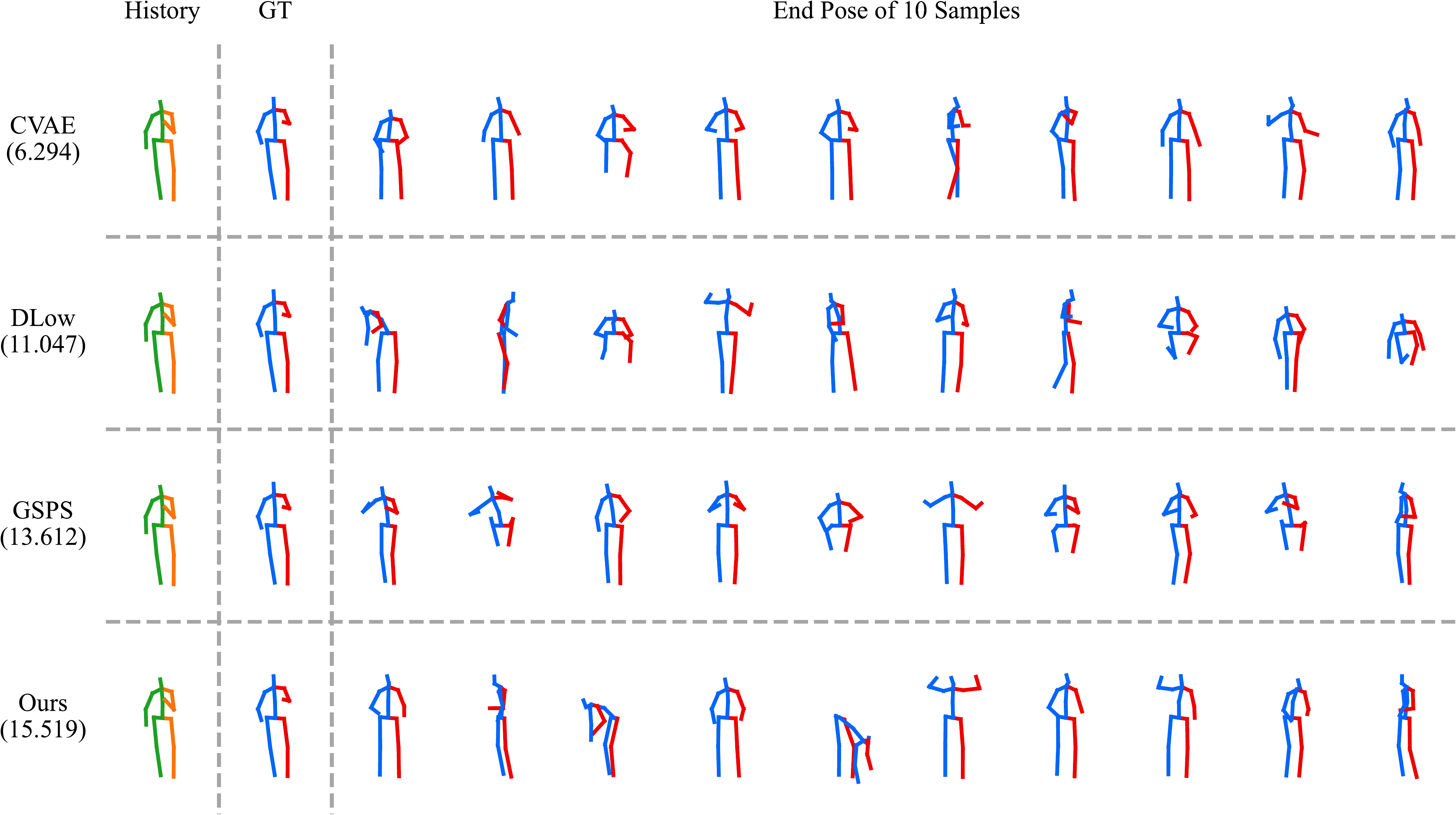}\\
		\includegraphics[width=0.49\linewidth]{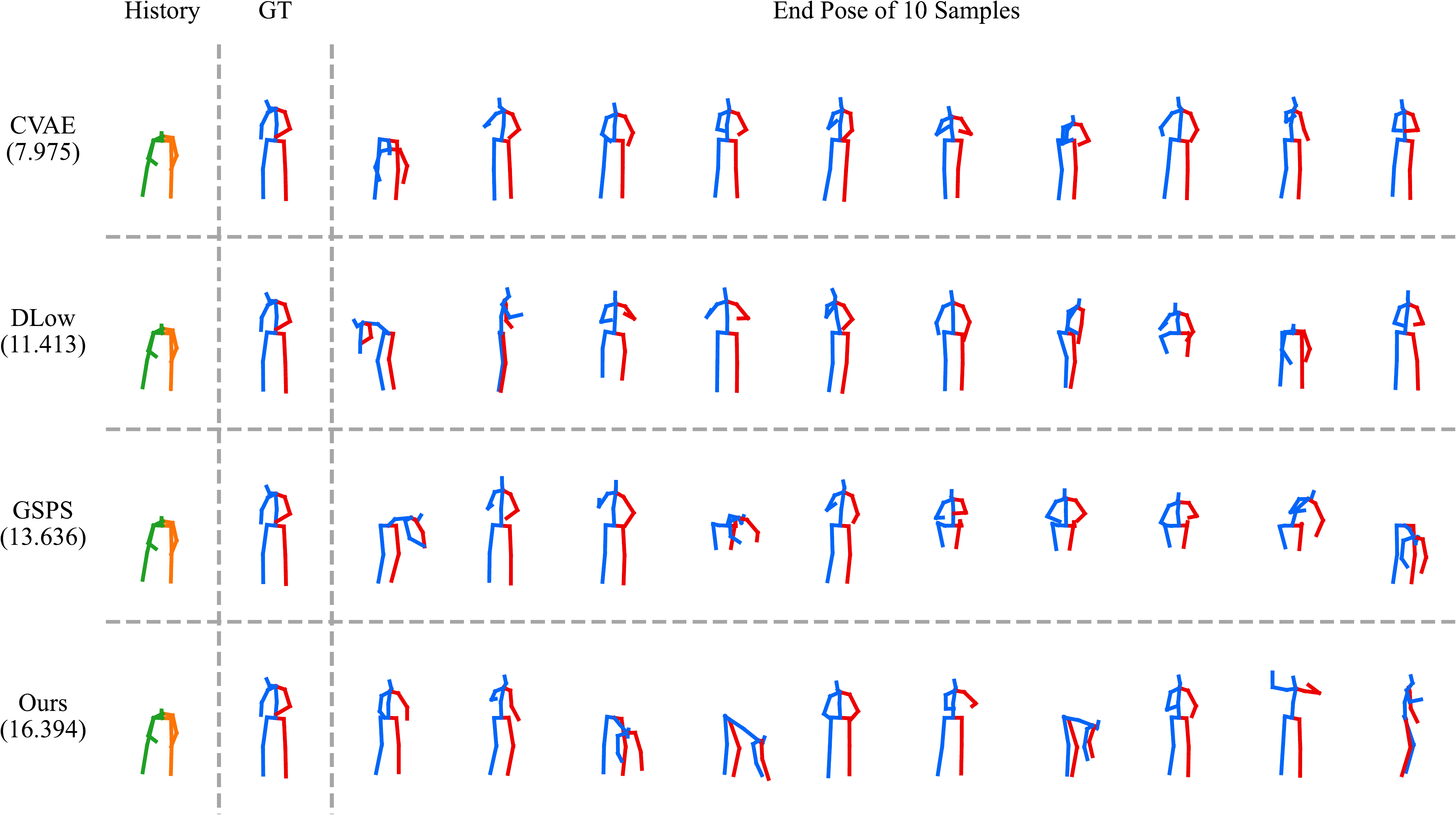}
		\includegraphics[width=0.49\linewidth]{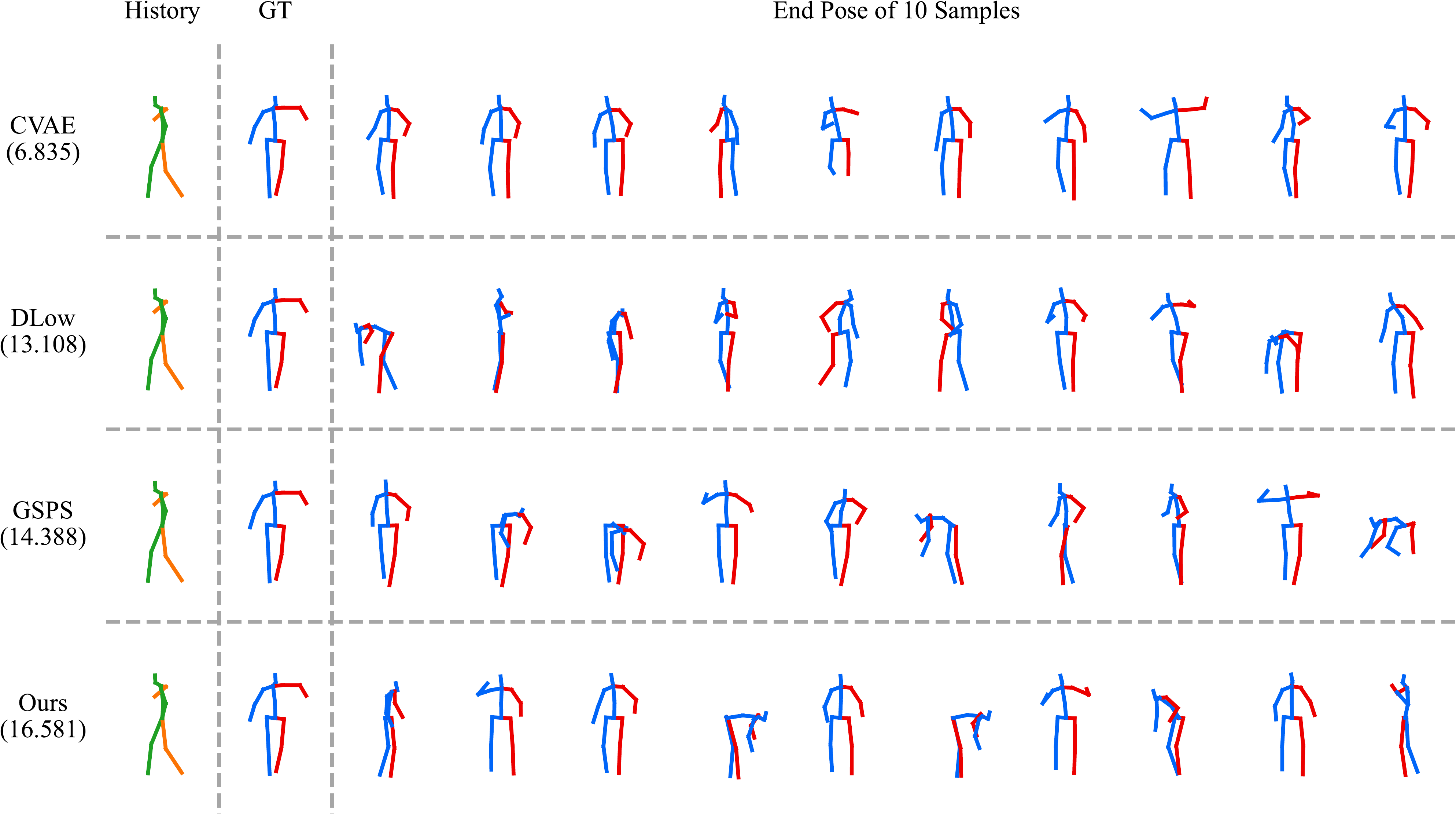} \\
		(a) Human3.6M \\
		\includegraphics[width=0.49\linewidth]{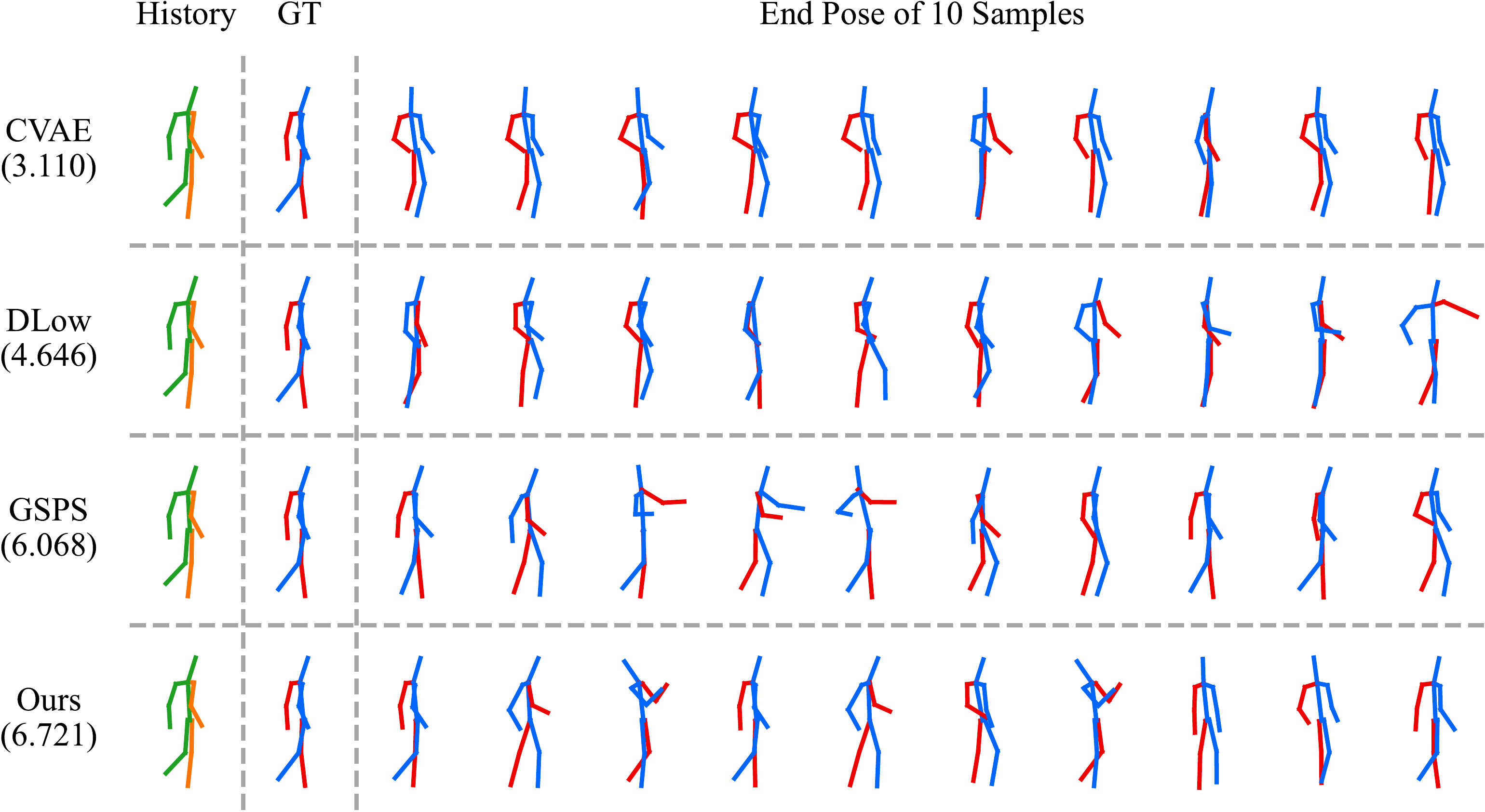}
		\includegraphics[width=0.49\linewidth]{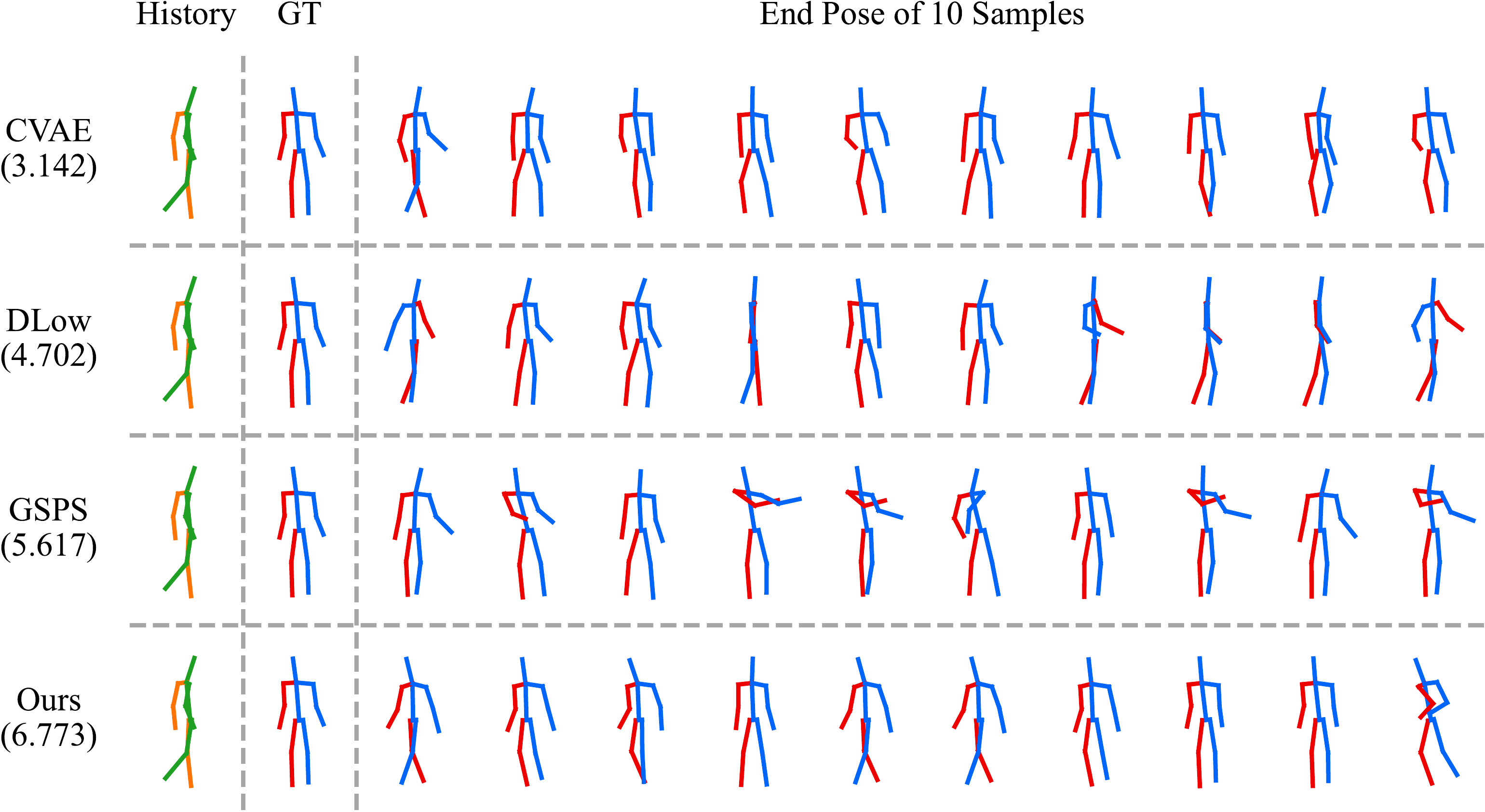} \\
		\includegraphics[width=0.49\linewidth]{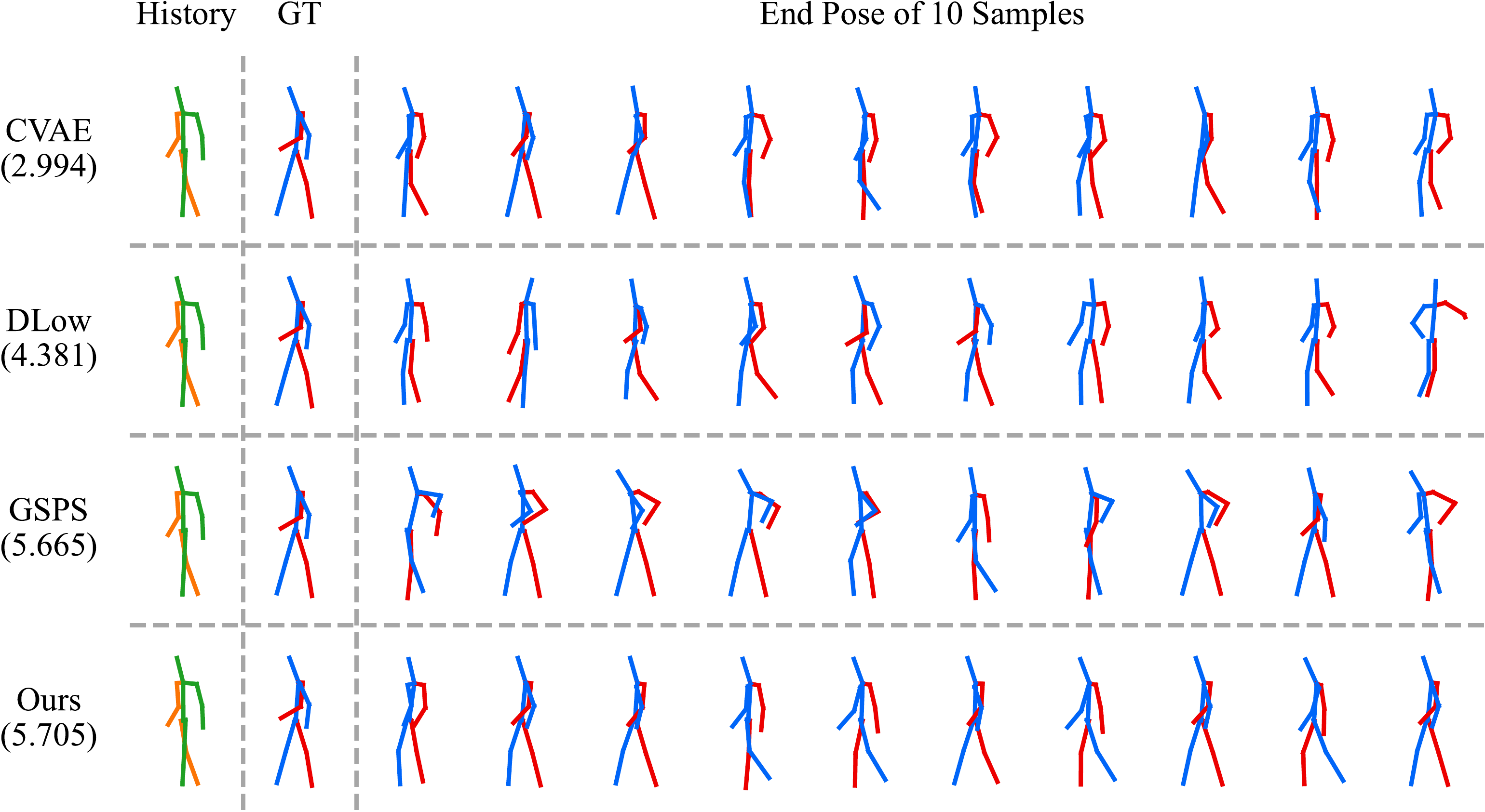}
		\includegraphics[width=0.49\linewidth]{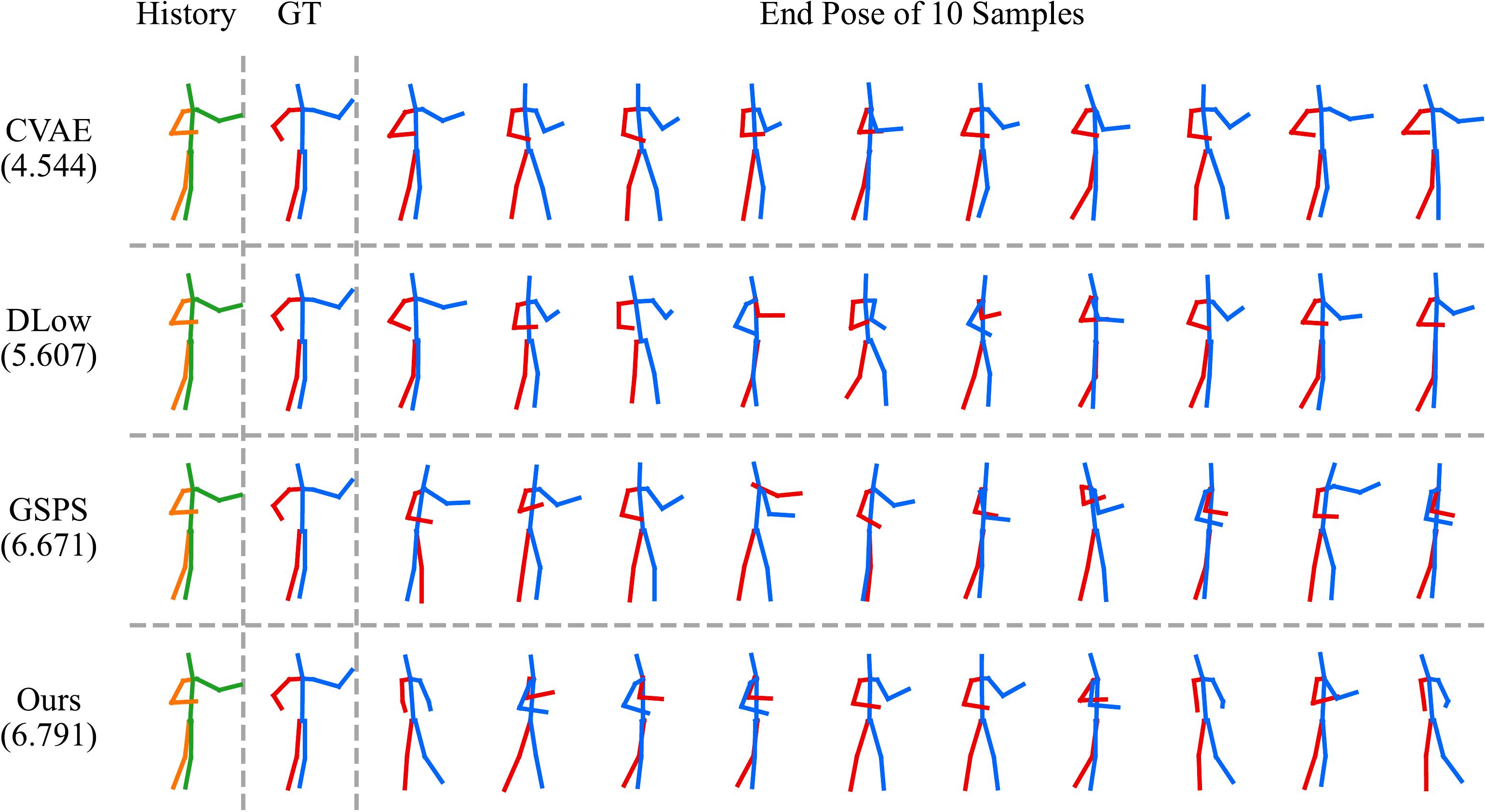}\\ 
		(b) HumanEva-I \\
		\caption{More qualitative results of CVAE, DLow, GSPS, and our method. The numbers in the brackets below the names of different methods show the diversity of the results computed by these methods. In these examples, our results are more diverse than the results of the other methods.}
		\label{fig:more-qual_2}
	\end{figure*}

	In Figure~\ref{fig:more-qual_1} and Figure~\ref{fig:more-qual_2}, we show more qualitative comparisons between CVAE, DLow \cite{yuan2020dlow}, GSPS \cite{mao2021generating} and our method on the Human3.6M dataset \cite{ionescu2013human3} and the HumanEva-I dataset \cite{sigal2010humaneva}. For each input sequence, we generate 50 future pose sequences by these methods, and show the end poses of ten of them. In the brackets under the names of different methods, we show the diversity of the corresponding results computed by these methods. For these examples, our method produces more diverse results than the other compared methods.

	
	\section{Failure Cases}
	
	\begin{figure*}[!t]
		\centering
		\includegraphics[width=0.49\linewidth]{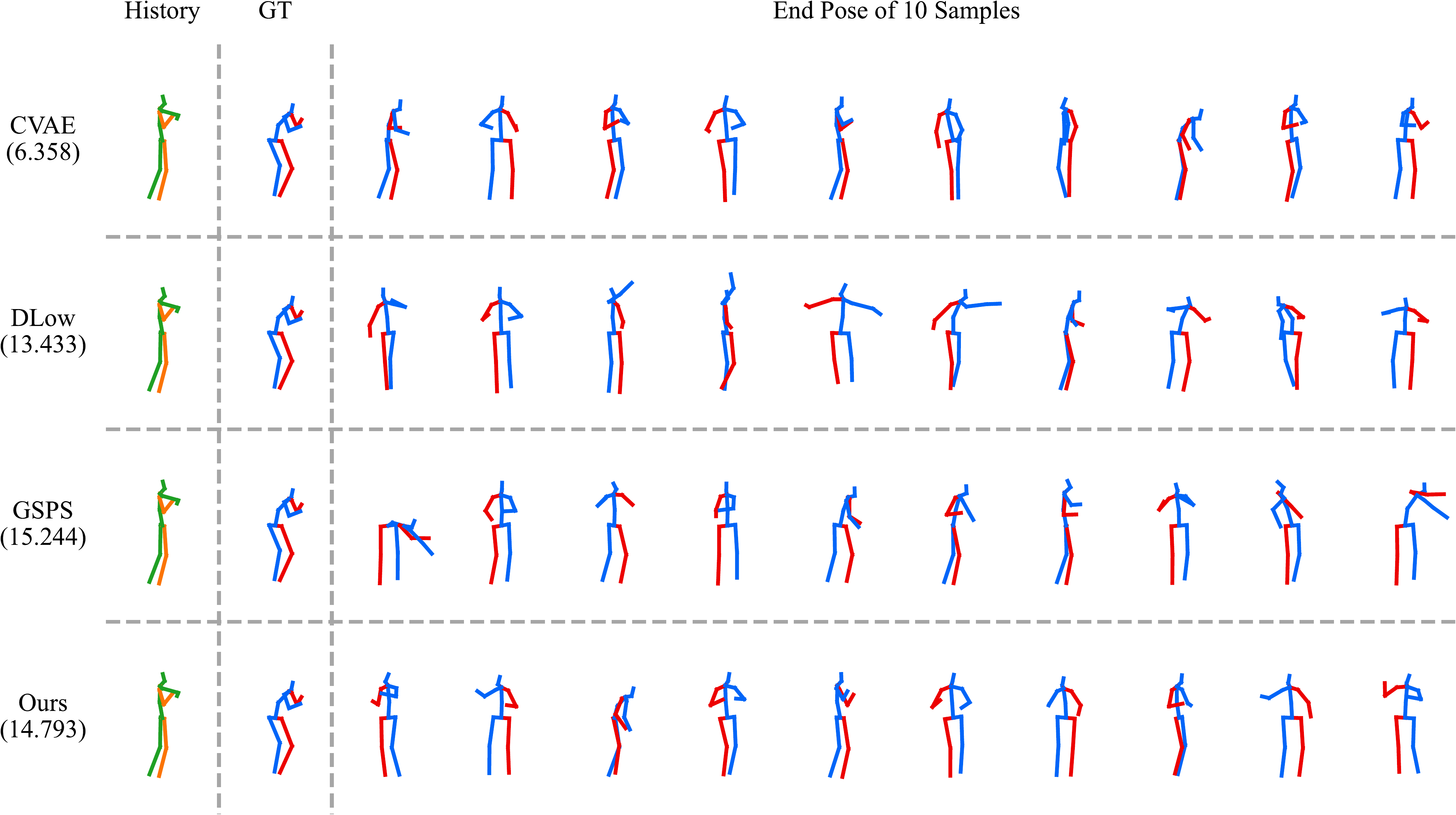} 
		\includegraphics[width=0.49\linewidth]{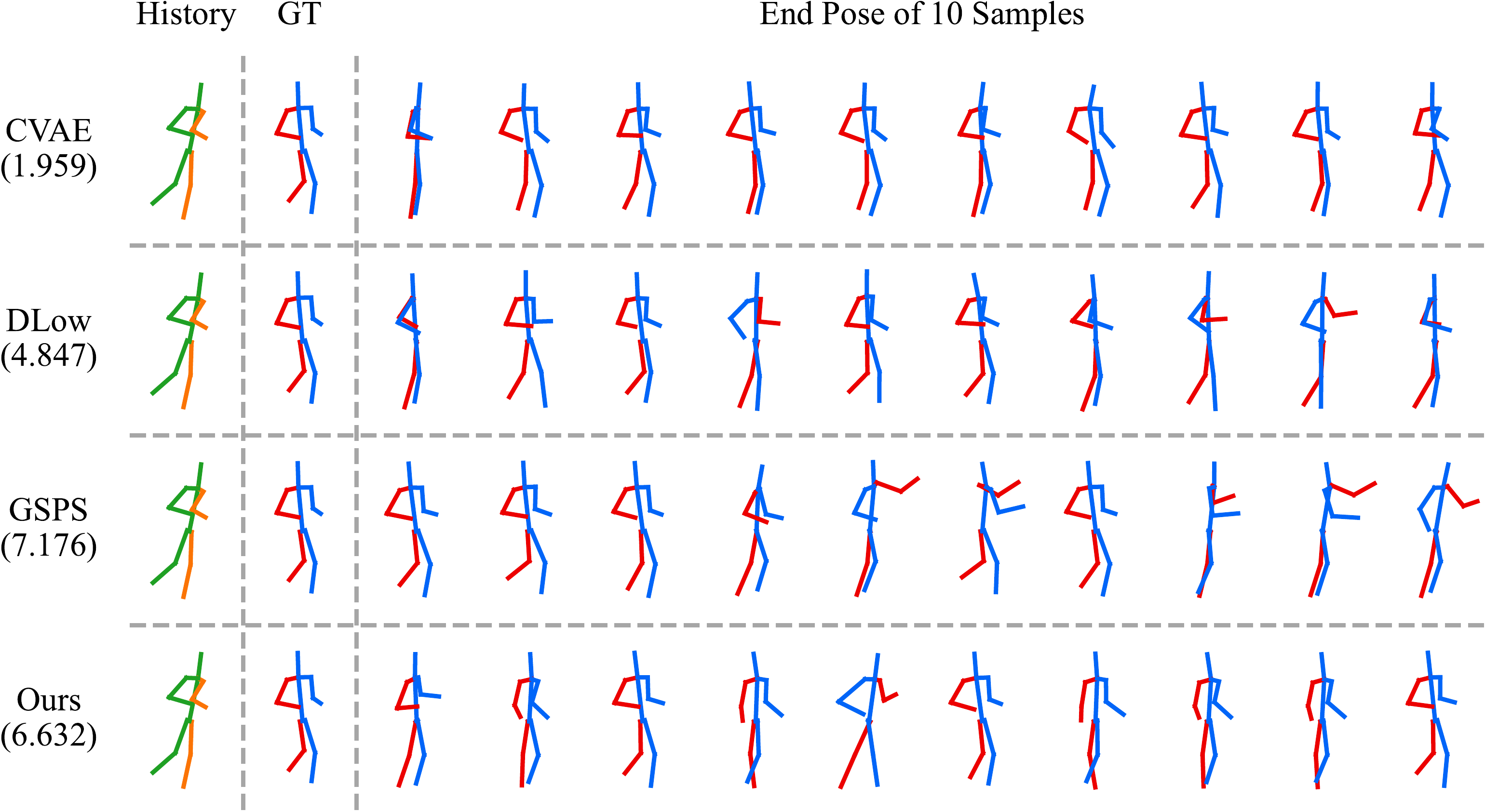}
		\flushleft \hspace{4cm} (a) Human3.6M \hspace{6.5cm} (b) HumanEva-I 
		\caption{Two examples for which our method generates lower diverse results than GSPS.}
		\label{fig:lower-diversity}
	\end{figure*}
	
	\begin{figure*}[!t]
		\centering
		\includegraphics[width=0.49\linewidth]{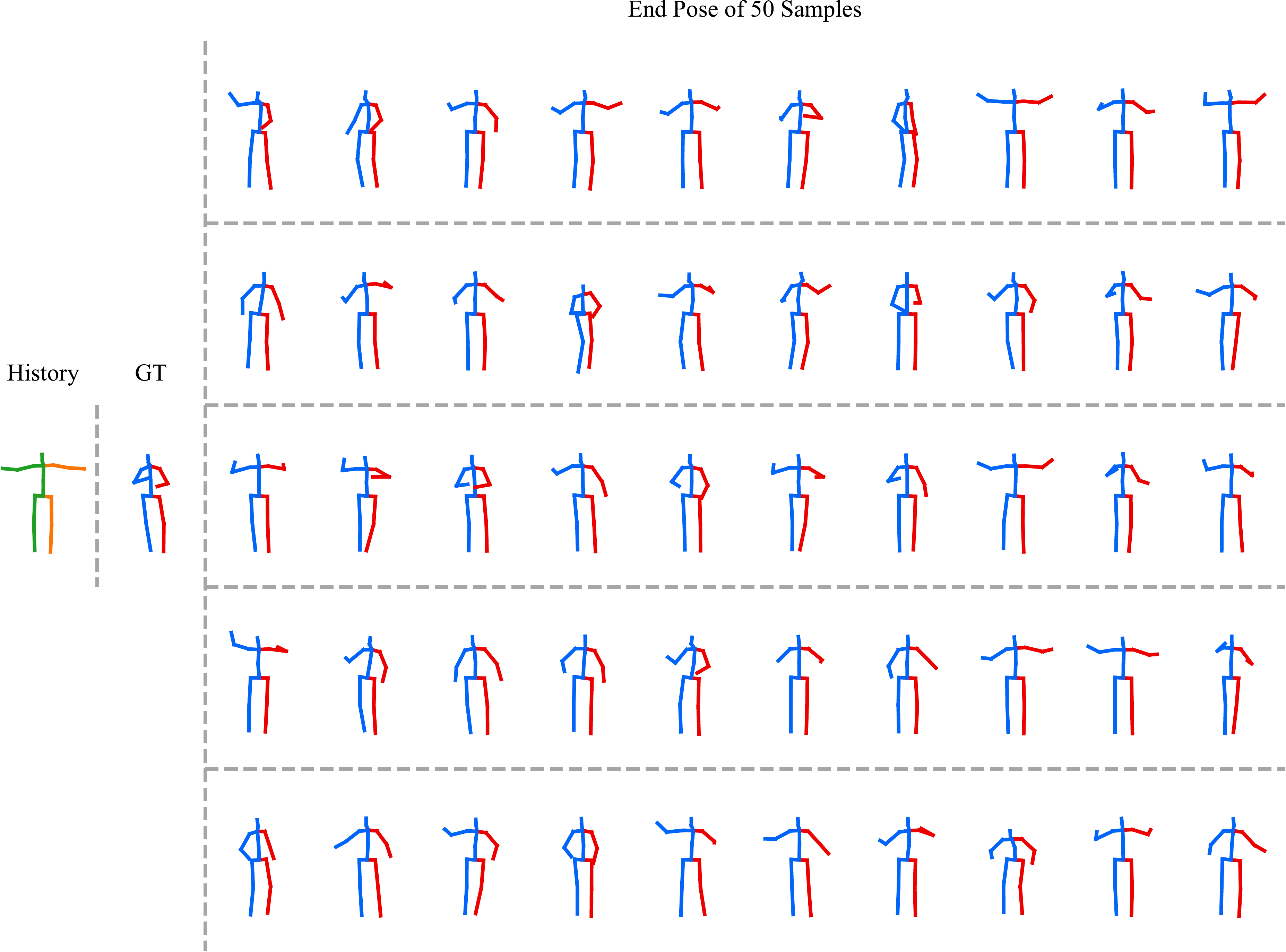} 
		\includegraphics[width=0.49\linewidth]{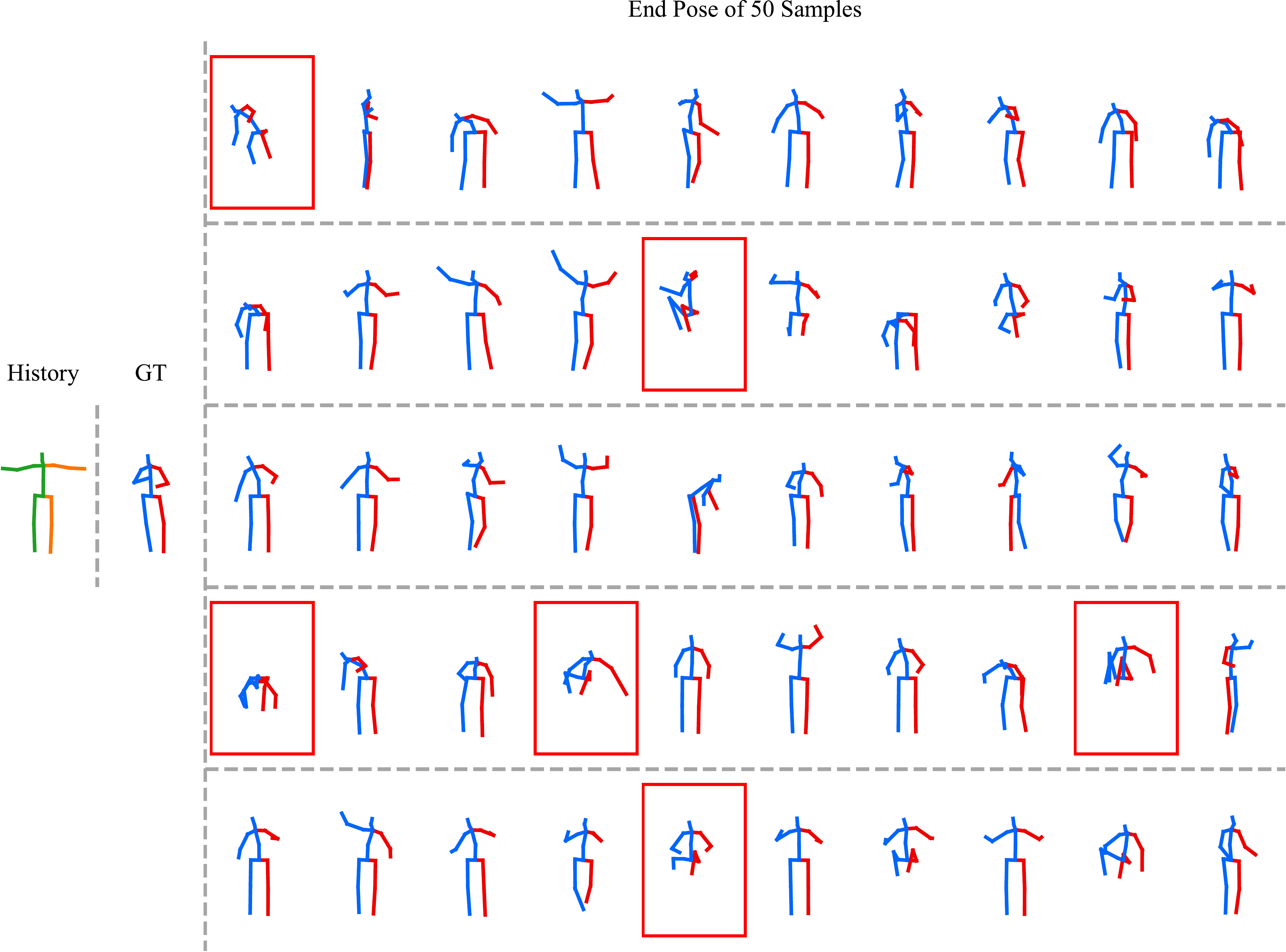} 
		\flushleft \hspace{4.5cm} (a) CVAE \hspace{7.8cm} (b) DLow 
		\includegraphics[width=0.49\linewidth]{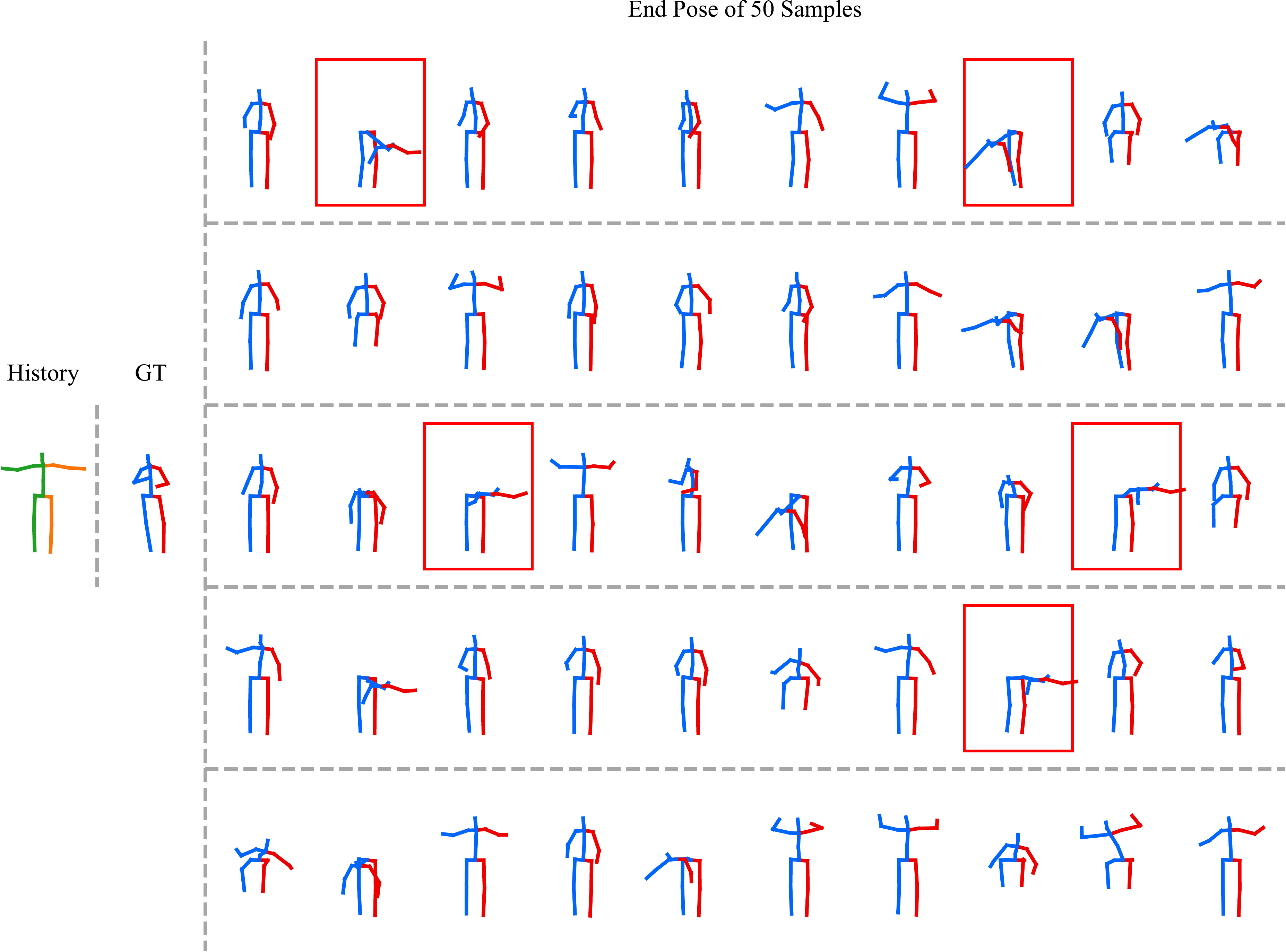} 
		\includegraphics[width=0.49\linewidth]{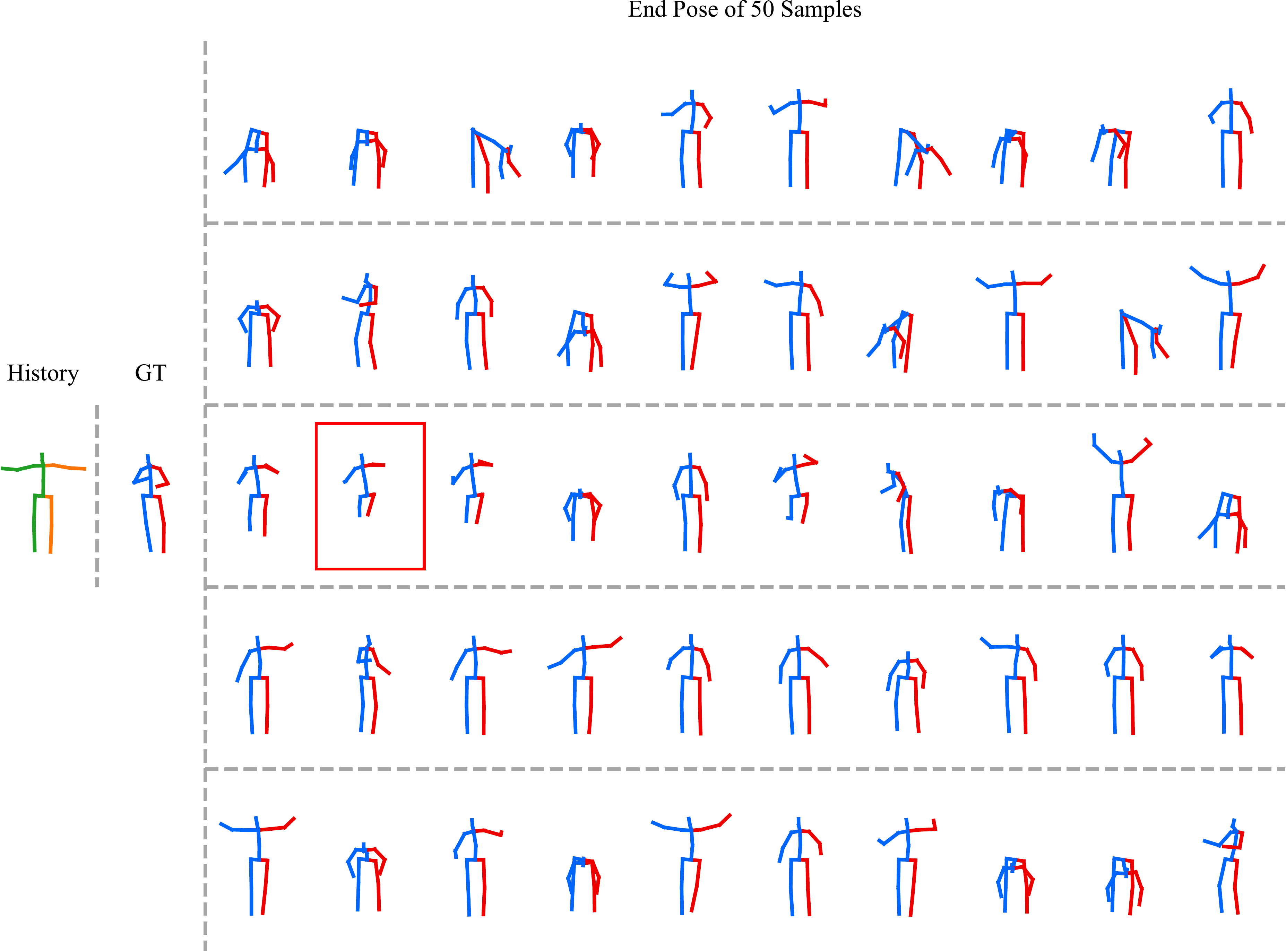} 
		\flushleft \hspace{4.5cm} (c) GSPS \hspace{7.8cm} (d) Ours 
		\caption{Examples of implausible poses in the results of DLow, GSPS, and our method.}
		\label{fig:invalid_poses}
	\end{figure*}
	
	Overall, our method is better than GSPS in term of diversity. Therefore, for most of the test cases our method generates more diverse results than GSPS. Inevitably there are cases for which our method generates less diverse results than GSPS. For example, Figure~\ref{fig:lower-diversity} shows such two cases. 
	
	While most of our results look reasonable, there are occasional ones that are implausible. Figure~\ref{fig:invalid_poses} shows some examples. We observe that DLow and GSPS suffer from this problem too, and the implausible poses in their results are highlighted too. One can reduce the number of implausible poses by toning down the requirement for diversity. A more effective way is to collect more data covering more actions of humans to enrich the training dataset.

\end{document}